\documentclass[lettersize,journal]{IEEEtran}
\usepackage{amsmath,amsfonts}
\usepackage{algorithmic}
\usepackage{algorithm}
\usepackage{array}
\usepackage[caption=false,font=normalsize,labelfont=sf,textfont=sf]{subfig}
\usepackage{textcomp}
\usepackage{stfloats}
\usepackage{url}
\usepackage{verbatim}
\usepackage{graphicx}
\usepackage{cite}
\usepackage{makecell}
\usepackage{booktabs}
\usepackage{multirow}
\usepackage{tabularx}
\usepackage{xcolor}
\colorlet{red}{black}
\begin{document}
% \linenumbers

\title{Cross-Modal UAV Object Tracking: State-Aware Representation Learning and A Unified Benchmark}

\author{Yun Xiao, Zhihong Hong, Jiandong Jin, Chenglong Li, Jin Tang, Amir Hussain

\thanks{
This research is partly supported by the National Natural Science Foundation of China (No. 62376004), the Natural Science Foundation of Anhui Province (2408085MF153). 
(Corresponding author: Jin Tang)

Yun Xiao, Zhihong Hong and Chenglong Li are affiliated with the Anhui Provincial Key Laboratory of Security Artificial Intelligence, School of Artificial Intelligence, Anhui University, Hefei 230601, China (e-mail: xiaoyun@ahu.edu.cn; hertzhong@foxmail.com; lcl1314@foxmail.com).

Jiandong Jin and Jin Tang are affiliated with the Anhui Provincial Key Laboratory of Multimodal Cognitive Computation, School of Computer Science and Technology, Anhui University, Hefei 230601, China (e-mail: jdjinahu@foxmail.com; tangjin@ahu.edu.cn)

Amir Hussain is affiliated with School of Computing, Engineering and the Built Environment, Edinburgh Napier University, Edinburgh EH10 5DT Scotland, UK (e-mail: A.hussain@napier.ac.uk)
}}

\markboth{Journal of \LaTeX\ Class Files,~Vol.~14, No.~8, August~2021}%
{Cross-Modal UAV Object Tracking: State-Aware Representation Learning and A Unified Benchmark}

\maketitle

\begin{abstract}
Unmanned Aerial Vehicle (UAV) object tracking has emerged as a popular research field with broad practical applications. Modern UAVs are increasingly equipped with both visible light and thermal infrared sensors. However, due to constraints in communication bandwidth, computational resources and power consumption, current systems often activate one modality and switch between modalities to maintain robust tracking in complex scenarios.
Such modality switch inevitably leads to significant appearance change and sudden spatial shift, posing great challenges for existing tracking algorithms.
To handle this problem, we propose a novel State-Aware Representation Learning Approach called SARLA, which perceives the inconsistent modality states of current frame with template and last frame in the target representations to adapt to the sudden changes in both appearance and position, for robust cross-modal object tracking.
In particular, we propose the Modality State Aware Representation Module (MSARM) and Spatial State Aware Representation Module (SSARM). MSARM guides the model to learn appearance correlation, bridging the modality gap, while SSARM models cross-frame spatial correlation to mitigate sudden spatial shift impacts.
In addition, we design a spatial shift prediction loss to further handle the effects of spatial variation caused by modality switch.
To promote the development of this research field, we establish a large-scale video benchmark called CM-UOT, which consists of 1079 cross-modal sequences with an average video length greater than 621 frames and encompasses over 671K frames in total.
Extensive experiments on CM-UOT dataset demonstrate the superior performance of the proposed SARLA against 20 excellent tracking methods.
The source code, datasets, and evaluation protocols associated with this work are publicly available at: https://github.com/hongsmile365/sarla-.
\end{abstract}


\section{Introduction}
\begin{figure}[htbp]
\centering
\includegraphics[width=\linewidth]{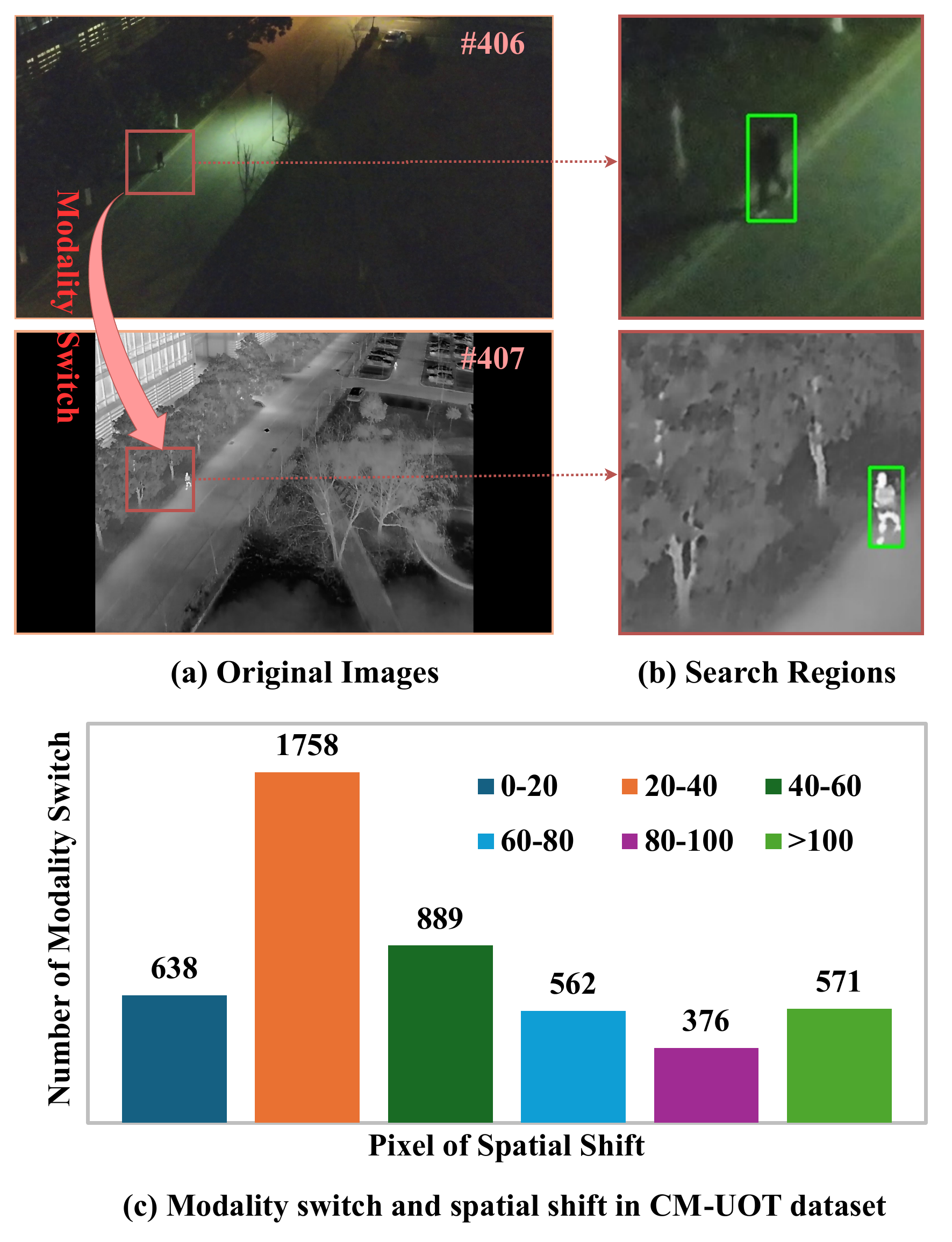}
\caption{Motivation demonstration. An example of modality switch shows the significant appearance change and spatial shift in (a) original images and (b) search regions. In (c), we count the occurrences of modality switch and spatial shift.
}\label{fig:motivation}
\end{figure}

\IEEEPARstart{U}{nmanned} Aerial Vehicle (UAV) object tracking is a fundamental task in computer vision, aimed at predicting the position of an object in subsequent frames using its known initial bounding box. It has various applications in fields such as emergency rescue, surveillance and monitoring, and has gained increasing attention in recent years.

Numerous works \cite{gtot7577747,li2017weighted,li2019rgb,li2020challenge,li2021lasher,lu2021rgbt} concentrate on enhancing the robustness of RGBT tracking by leveraging the complementary strengths of dual modalities under all-day and all-weather conditions. Following these foundational studies, advanced architectures are systematically explored \cite{liu2023quality, li2023multi, hui2023bridging} to address challenges such as feature misalignment and modality imbalance. Recent efforts \cite{liu2024rgbt,cao2024bi,tcsvt10614652,lu2025duality,tcsvt10949219} further enhance the robustness and adaptability of RGBT trackers in highly dynamic and adverse conditions.

However, due to restrictions in communication bandwidth, computational resources and battery drain, the simultaneous acquisition of data from both modalities in current UAV systems is unachievable. These systems often activate one modality and switch between modalities for real-time tracking.
Hence, research on the cross-modal UAV object tracking task carries significant practical value.

At present, some works explore RGB and near-infrared (NIR) cross-modal object tracking task in general scenarios.
To bridge the modality gap,  MArMOT~\cite{li2022cross} introduces a modality-aware feature learning framework through a three-stage training strategy, and MAFNet~\cite{liu2024cross} develops two modality-specific branches combined with an adaptive weighting module to achieve effective cross-modal feature fusion.
Nevertheless, there is an essential difference between RGB-NIR cross-modal object tracking task and RGBT cross-modal object tracking task in UAV-based scenarios, as the latter often alternates between two modality-specific cameras for tracking. Due to the positional differences between cameras and the complexity of UAV movement patterns, which lead to significant spatial shift, present models fail to account for the impact of spatial shift, which can lead to target loss, thereby limiting their effectiveness in UAV scenarios. Additionally, 
the appearance gap between RGB and TIR modalities is often more pronounced, which makes it difficult for present models to adapt to the RGBT cross-modal tracking task.

As shown in Fig.~\ref{fig:motivation} (a) and (b), the issue of modality switch inevitably leads to significant appearance change and sudden spatial shift in the cross-modal UAV object tracking task. 
The spatial shift suggests the entire image transforms, affecting both the target object and the background.
On the one hand, the shift from RGB to TIR imaging entails a significant loss of chromatic information and textural details, posing a substantial challenge for cross-modal UAV object tracking. 
On the other hand, the dual-camera system used in UAVs exhibits inherent positional discrepancies and diverse motion patterns, making it difficult to predict the direction and scale of spatial shift following modality switch.
Fig.~\ref{fig:motivation} (c) demonstrates that significant appearance change and sudden spatial shift are prevalent in the cross-modal UAV object tracking task and necessitate the implementation of appropriate countermeasures.

To address the challenges of modality switch and spatial shift, we aim to guide the model by perceiving the changes in modality and spatial variations, helping the model establish the relationship between cross-modal target appearances and spatial locations. This approach mitigates the impact of significant appearance change and complex spatial shift in cross-modal UAV object tracking. Furthermore, we construct the first RGBT cross-modal UAV object tracking video benchmark dataset, providing a research foundation for the development of RGBT cross-modal UAV object tracking algorithms.

First, we propose a State-Aware Representation Learning Approach (SARLA) for adaptive cross-modal UAV object tracking. To reduce the computational burden and meet the efficiency requirements of UAV-based edge devices, our approach employs a single-branch structure to directly process cross-modal data, unlike previous methods rely on multiple modality-specific branches. To mitigate cross-modal appearance discrepancies, we design a modality state-aware representation module (MSARM), which introduces a modality state token to guide the model in recognizing modality changes and capturing cross-modal appearance correlations.

Furthermore, since UAV-based scenarios exhibit more pronounced spatial shift during modality transitions, we propose a spatial state-aware representation module (SSARM), which introduces a novel learnable spatial state token and samples an additional reference region from the previous frame to capture spatial state variations by linking spatial information from the reference and search regions. Due to the accumulation of errors during the tracking process, the reference frame may not always provide reliable target information. To stably model the cross-modal target appearance relationship, we restrict MSARM to operate between the initial template and the current frame. Specifically, we employ an attention mask mechanism to enable the two state tokens to effectively capture modality and spatial states, and allow for parallel processing of all tokens, avoiding redundant multi-branch designs.

Second, to provide a research benchmark for RGBT cross-modal tracking in UAV scenarios and tackle the problem of the scarcity of cross-modal UAV object tracking data, we establish a large-scale video benchmark called CM-UOT, which consists of 1079 cross-modal sequences with an average video length greater than 621 frames and encompasses over 671K frames in total.
As shown in Fig.~\ref{fig:motivation}, our dataset exhibits spatial shift at different scales, and in each sequence, the number of frames with modality inconsistencies compared to the initial state occupies a significant proportion.

Unlike CMOTB~\cite{liu2024cross}, our dataset systematically accounts for the unique characteristics of cross-modal UAV object tracking tasks as follows.
(1) Our dataset employs a TIR sensor commonly equipped on UAVs, which exhibits significant differences in imaging principles compared to the NIR sensor used in CMOTB, often resulting in more pronounced appearance change. 
(2) Our dataset fully considers unique challenges in UAV scenarios, such as small targets, intense motion, camera rotation, and zooming. 
(3) Our dataset incorporates diverse spatial shift variations. Since UAVs typically use independent RGB and TIR sensors, modality transitions inherently introduce spatial shift. Moreover, the complex motion patterns of UAVs further exacerbate these variations.
(4) To effectively evaluate whether tracking methods can handle real-world UAV motion, we annotate 19 challenge attributes for each video and introduce five novel challenge attributes specifically designed for UAV scenarios.

\textcolor{red}{Different from simply appending learnable tokens to an existing one-stream tracker, SARLA is motivated by a state decomposition of modality switch in cross-modal UAV object tracking. We observe that modality switch induces two coupled but distinct challenges: cross-modal appearance variation and cross-frame spatial displacement, which are formulated as the appearance state and the spatial state, respectively. Based on this formulation, SARLA introduces two dedicated state-aware representation modules. The modality-state token and spatial-state token are not generic auxiliary tokens, but are explicitly guided by an attention-mask mechanism to focus on two different state-reasoning sub-tasks. Furthermore, modality-state classification and spatial-shift prediction are introduced to supervise the two state tokens, enabling the model to learn interpretable state-aware representations for handling modality switch.}

The main contributions of this work are summarized as follows:
\begin{itemize}
\item We introduce a new task, namely cross-modal UAV object tracking, which stems from the selective modality activation mechanism of UAVs with dual modality. Such a task is of great significance in research and has substantial practical value.  
\item We propose a novel State-Aware Representation Learning Approach for robust cross-modal UAV object tracking, namely SARLA, which model the cross-modality appearance correlation guided by the modality state token and learn the cross-frame spatial correlation guided by the spatial state token.
\item We establish a large-scale unified video benchmark for cross-modal UAV object tracking called CM-UOT.
It offers a strong foundation for advancing cross-modal UAV object tracking research.
We also provide an aligned version of CM-UOT to support focused research.
\item 
We conduct a comprehensive evaluation and analysis of different tracking algorithms on CM-UOT dataset.
Extensive experiments demonstrate that our method achieves superior performances compared to other trackers.
\end{itemize}

\section{Related Work}
\subsection{Visual Object Tracking}
One crucial factor impacting the robustness of visual object trackers is the variation in target appearance. To tackle this problem, numerous approaches have emerged.
GRM~\cite{gao2023generalized} utilizes an adaptive token division strategy to alleviate target-background confusion.
Moreover, some studies attempt to leverage the spatiotemporal information in video streams to handle the variations. SeqTrack~\cite{chen2023seqtrack} employs a simple encoder-decoder transformer architecture that transforms the four-dimensional bounding box coordinates into a sequence of discrete tokens, generating predictions in an autoregressive fashion. 
ARTrackV2~\cite{bai2024artrackv2} employs a pure encoder architecture and incorporates appearance tokens, achieving superior performance with enhanced efficiency. 
LGTrack~\cite{TCSVTlgtrack} incorporates a dynamic template update mechanism to adapt to appearance change during long-term tracking.

Although these methods can effectively adapt to dynamic appearance variation over time, they struggle to handle 
significant appearance change across different modalities in UAV object tracking. The appearance transformation caused by modality gaps is not only due to changes in viewpoint and scale but also accompanied by significant changes in color and detail information. Additionally, these methods are prone to tracking drift when there is severe spatial displacement between consecutive frames.

\subsection{Multi-modal Object Tracking}
In the multi-modal (RGBT) object tracking task, the modality gap poses an urgent problem that demands attention. Recently, numerous works~\cite{RGB-TTracking11112641,TCSVTrt10973112,TCSVTrt11112670} have emerged to tackle this issue.
TBSI~\cite{hui2023bridging} uses templates to bridge cross-modal interaction between RGB and TIR regions and updates templates with enriched contexts.
CKD~\cite{lu2024breaking} proposes a coupled knowledge distillation framework with two student networks and style distillation loss. 
% It distills content knowledge from RGB and TIR teacher networks into students via feature decoupling to eliminate the gap.
BAT~\cite{cao2024bi} presents a multi-modal visual prompt tracking model with a bi-directional adapter for cross-prompting and a light feature adapter for adaptive modality information transfer and fusion. To address the modality-missing challenge, IPL~\cite{Lu2024} designs the invertible prompter by incorporating the full reconstruction of the input available modality from the generated prompt.

Multi-modal (RGBT) object tracking features dual-modal data at each frame, including the initial frame. In contrast, cross-modal (RGBT) UAV tracking has only one modality's data per frame and an initial frame in one modality, which can be regarded as a multi-modal task where one modality is consistently absent.
On the one hand, running RGBT trackers on our dataset CM-UOT, such as through copying and generating, is a feasible approach, but it will double the computational burden. 
On the other hand, our benchmark CM-UOT takes into account the limitations and attributes of UAVs. In contrast, VTUAV \cite{zhang2022visible}, a multi-modal UAV object tracking benchmark, fails to do so.

\subsection{Cross-modal Object Tracking}
Based on the characteristics of surveillance cameras, which switch between RGB and NIR imaging in accordance with light intensity, MArMOT~\cite{li2022cross} introduced the first cross-modal tracking benchmark, namely CMOTB, along with a three-stage modality-aware learning framework. The three-stage training strategy is specifically devised for the backbone, the modality-aware branch, and the ensemble module, respectively, with the aim of achieving a more effective target representation.
Later, to handle the challenge of modality adaptation, an extended version of CMOTB~\cite{liu2024cross} is released, and a cross-modal feature fusion approach is proposed. This approach incorporates two modality-specific branches that are combined with an adaptive weighting module, enabling the effective fusion of features from different modalities.
Furthermore, ProroTrack~\cite{LIU2025102941} utilizes multi-modal prototypes with a complex prototype update strategy to better capture the complex patterns and dependencies across time frames and modalities.

The cross-modal object tracking task in UAV scenarios differs significantly from that in general scenarios. The differences lie in aspects such as the type of modality switch, unique attributes like tiny targets and diverse motion patterns, and the presence of spatial shift.
On the one hand, although there are cross-modal tracking benchmarks for general scenes~\cite{li2022cross,liu2024cross}, the differences between the two tasks limit their applicability in evaluating UAV cross-modal tracking algorithms. To promote the development of cross-modal UAV object tracking, we have established a large-scale video benchmark named CM-UOT. On the other hand, these algorithms do not address the challenge of spatial displacement and are unable to solve the issues in our task. Furthermore, they adopt redundant and modality-specific branch designs. In contrast, our method uses a single-branch architecture and does not rely on complex design updates, enabling it to effectively address the challenges of appearance variation and spatial displacement caused by modality switch.

\begin{figure*}[htbp]
\centering
\includegraphics[width=0.9\linewidth]{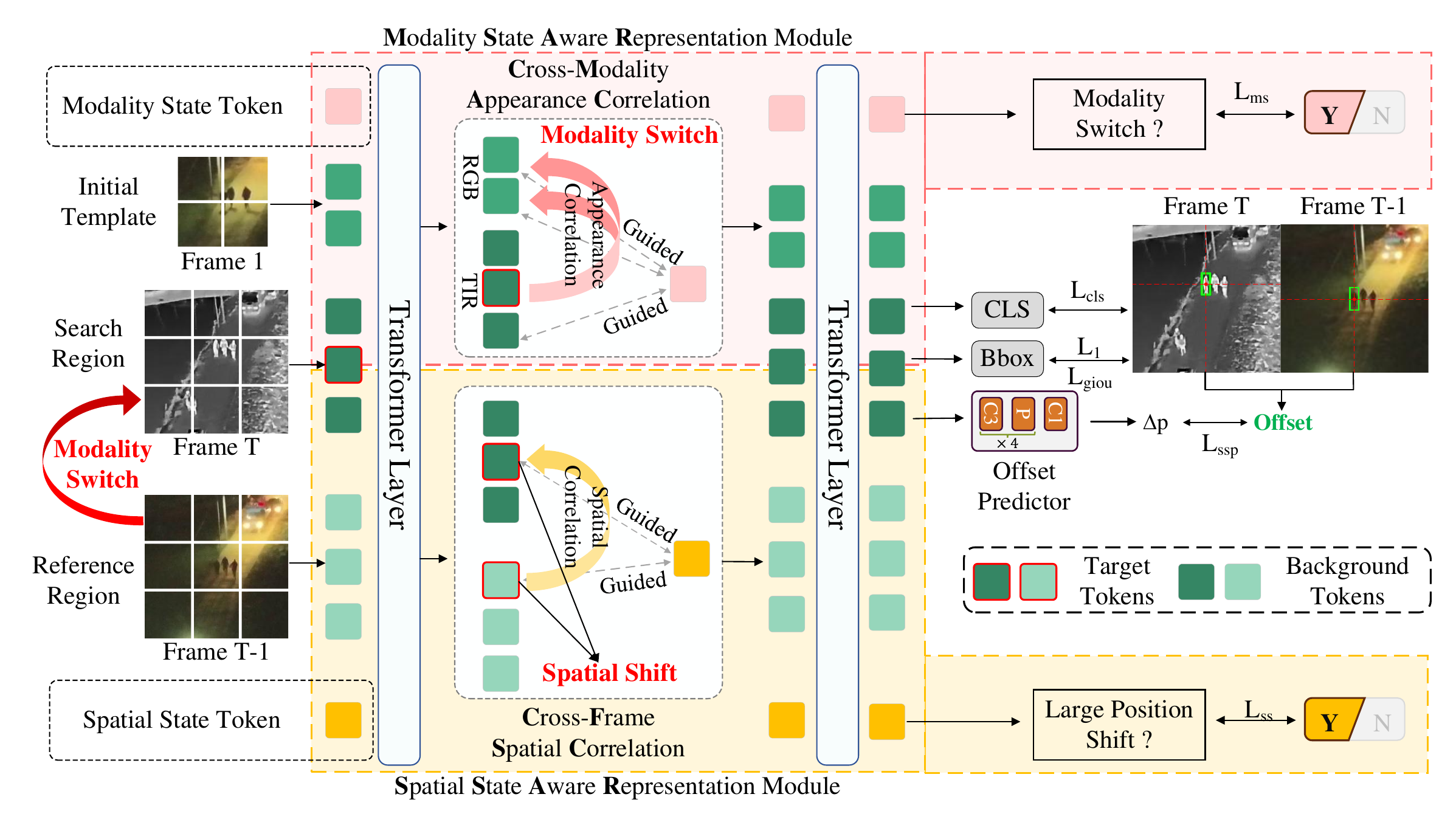}
\caption{The overall framework of our proposed method is illustrated as follows. We utilize a Transformer encoder to process all tokens in parallel. Subsequently, search region tokens are sent to the class head, the bounding box head, and the spatial shift head for predicting the class, bounding box, and spatial shift, respectively. The modality state token and spatial state token are used to predict their respective states.
C3: Convolutional layer with a $3\times 3$ kernel; P: Max-pooling layer with a $2\times 2$ window; C1: Convolutional layer with a $1\times 1$ kernel.
}
\label{fig:method}
\end{figure*}

\section{Methodology}
In this section, we introduce our proposed method, the State-Aware Representation Learning Approach (SARLA). 
It consists of two main modules: the Modality State Aware Representation Module (MSARM) and the Spatial State Aware Representation Module (SSARM).
First, MSARM is designed to guide the model in learning the appearance correlation, thereby bridging the modality gap. 
Additionally, in order to mitigate the impact of spatial shift, SSARM focuses on modeling the cross-frame spatial correlation.
A novel loss function constrains the aforementioned learning process, and this entire process functions in an end-to-end manner.

\subsection{Overall Architecture}
In cross-modal UAV object tracking, modality switch inevitably induces significant appearance change and sudden spatial shift. 
The spatial shift indicate that the entire image undergoes changes, encompassing both the target object and the background.
The conventional methods~\cite{li2022cross, liu2024cross, LIU2025102941}, partially tackle these issues to a certain extent by taking significant appearance change into account through a two-branch network or a complex update strategy. 
However, these methods fail to account for the spatial shift challenges and encounter difficulties in tracking the target when there are large positional shift.
When a modality switch occurs, these limitations often lead to incorrect tracking results. 
Due to the recursive nature of tracking frameworks, errors introduced during modality switch persist until the end, which is a critical factor contributing to their poor performance.

To address the issues mentioned above, we introduce the State-Aware Representation Learning Approach (SARLA), as illustrated in Fig.~\ref{fig:method}. This approach is trained in an end-to-end fashion.
The input to SARLA consists of a tuple of images: the template image $I_{z}$, the reference region $I_{ref}$, and the search region $I_{x}$, where each is represented as $\mathbb{R}^{3 \times H_t \times W_t}$, $t = z, ref, x$, with each image modality being either visible light (VIS) or thermal infrared (TIR). These images are divided into patches and flattened into sequences. Subsequently, they are mapped to a sequence of tokens $\mathbf{E}_t \in \mathbb{R}^{N_t \times D}$ through a learnable linear projection layer, where $N_t$ refers to the length of the tokens of each image, and $D$ is the dimensional of tokens. This process is represented as follows.
\begin{equation}
\mathbf{E}_t = \theta_p (patchify(I_t))
\end{equation}
where $patchify(\cdot)$ represents a series of operations, including dividing the image into patches and flattening them, and $\theta_p(\cdot)$ represents the linear projection layer.
To indicate the positional information between tokens and distinguish between the template, reference region, and search region, we introduce a learnable positional encoding $\mathbf{P}_t \in \mathbb{R}^{N_t \times D}$ for each image. We add the positional encoding to the corresponding tokens of each part, resulting in the visual features $\mathbf{F}_t \in \mathbb{R}^{N_t \times D}$ for the model.
\begin{equation}
\mathbf{F}_t = \mathbf{E}_t + \mathbf{P}_t
\end{equation}

Meanwhile, the learnable modality state token \(\mathbf{T}_{ms} \in \mathbb{R}^{1 \times D}\) and the learnable spatial state token \(\mathbf{T}_{ss} \in \mathbb{R}^{1 \times D}\) are combined with learnable 1D position embeddings \(\mathbf{P}_{ms}\) and \(\mathbf{P}_{ss}\), respectively. As a result, the modality state token feature \( \mathbf{F}_{ms} \in \mathbb{R}^{1 \times D} \) and spatial state token feature \( \mathbf{F}_{ss} \in \mathbb{R}^{1 \times D} \) are generated.
\begin{equation}
 \mathbf{F}_{ms} = \mathbf{T}_{ms} + \mathbf{P}_{ms}, \mathbf{F}_{ss} = \mathbf{T}_{ss} + \mathbf{P}_{ss}.
\end{equation}

Subsequently, we concatenate the visual features $\mathbf{F}_t$ and state token features to form the input sequence $\mathbf{F}_u^0$ for the model. The resulting vector $\mathbf{F}_u^0$ is then fed into several Transformer encoder layers $L$ with an attention mask, as shown by the formula:

\begin{equation}
\mathbf{F}_u^{i + 1} = L^i(\mathbf{F}_u^i; \mathbf{A}),
\end{equation} 
where $\mathbf{A}$ denotes the attention mask and $i \in \{0, 1, 2, \ldots, 11\}$ is the index of transformer layer.

\begin{figure}[htbp]
\centering
\includegraphics[width=0.65\linewidth]{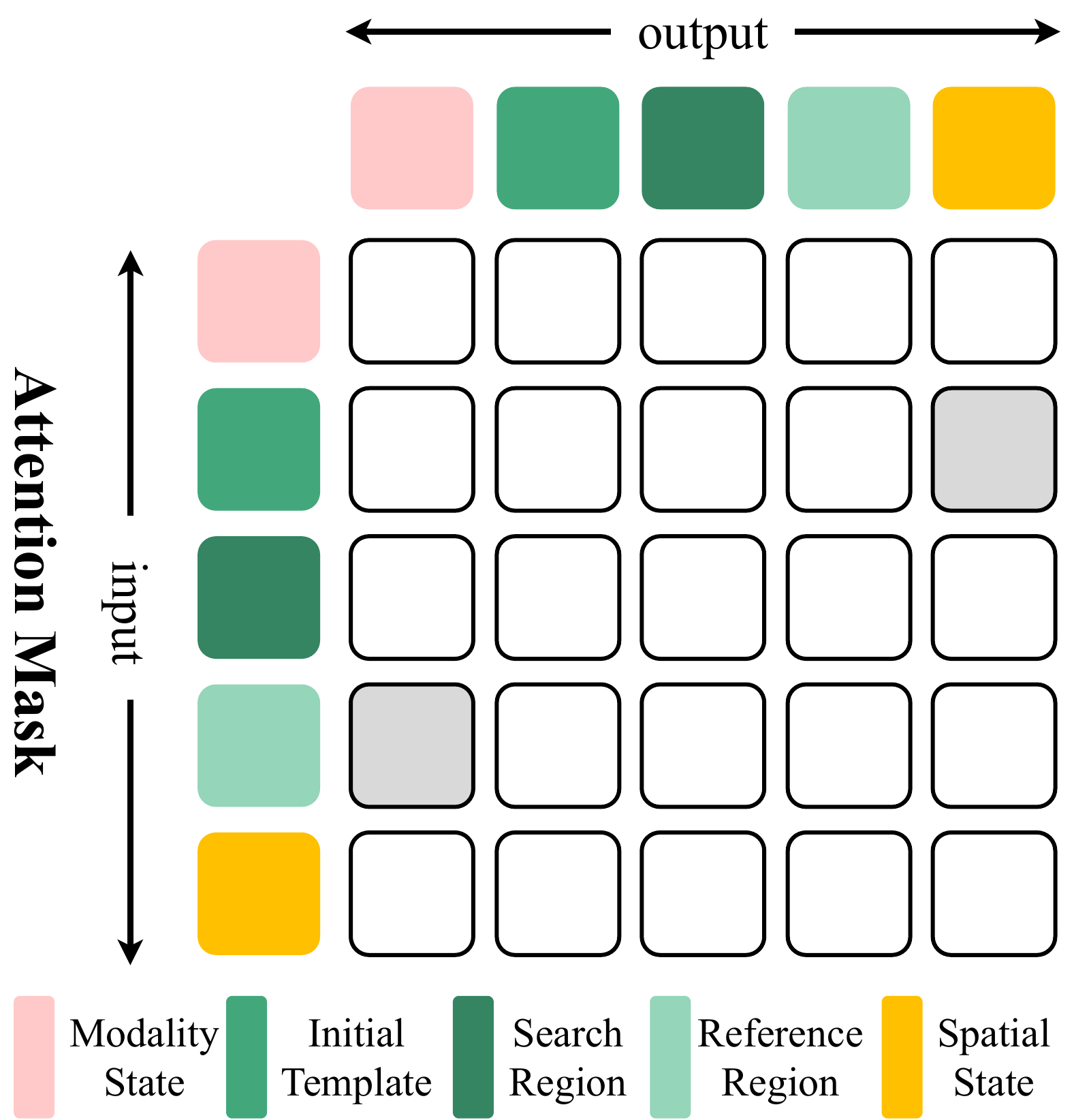}
\caption{\textcolor{red}{Illustration of the attention mask used in SARLA. The mask controls the interactions among the template, reference, search, modality-state token, and spatial-state token.}}
\label{fig:mask}
\end{figure}

\textcolor{red}{Specifically, $\mathbf{A}$ denotes a binary attention mask that explicitly controls token-to-token interactions, where $\mathbf{A}_{ij}=1$ allows the $i$-th token to attend to the $j$-th token, while $\mathbf{A}_{ij}=0$ blocks this attention connection. As illustrated in Fig.~\ref{fig:mask}, the white blocks indicate allowed attention interactions, while the gray blocks indicate suppressed interactions. In our design, the mask follows the functional roles of the two state tokens. The modality-state token $\mathbf{F}_{ms}$ is designed to model modality variation between the initial template and the current search region. Therefore, it is allowed to interact with the initial template features $\mathbf{F}_{z}$ and the current search features $\mathbf{F}_{x}$, while its interaction with the reference-frame features $\mathbf{F}_{ref}$ is masked out. In contrast, the spatial-state token $\mathbf{F}_{ss}$ is responsible for capturing spatial changes from the reference frame to the current search region; hence, it is allowed to attend to $\mathbf{F}_{ref}$ and $\mathbf{F}_{x}$, while its interaction with $\mathbf{F}_{z}$ is suppressed. This mask design decouples modality-state reasoning from spatial-state reasoning, thereby avoiding redundant or inconsistent information exchange between unrelated token groups.}

The attention masking mechanism ensures that modality state token $\mathbf{F}_{ms}$ specifically captures modality changes, and spatial state token $\mathbf{F}_{ss}$ perceives spatial shift.
Specifically, we restrict the interactions of \(\mathbf{F}_{ms}\) to only template features \(\mathbf{F}_{z}\) and search features \(\mathbf{F}_{x}\), and limit \(\mathbf{F}_{ss}\) to interact solely with 
reference features \(\mathbf{F}_{ref}\) and search features  \(\mathbf{F}_{x}\). This selective attention masking enforces the desired functional separation and allows the model to process the token sequences in parallel efficiently.

\subsection{Modality State Aware Representation Module}

To address the issue of significant appearance change, we propose Modality State Aware Representation Module (MSARM).
Unlike previous methods, which either employ a two-branch network or utilize a complex update strategy, MSARM adopts a single branch and incorporates a modality token to handle such appearance change. This helps reduce the complexity of the model, which is crucial for edge devices such as drones.
In particular, it employs a learnable modality state token to guide the module to learn the cross-modal appearance correlation.
The modality state interacts with the initial template $I_z$ that offers reliable object information and search region $I_x$ through the attention mechanism.
During the forward propagation process, changes in the modality state are gradually perceived. 

When an inconsistency occurs, the token separates the modality-related features from the target appearance features. Simultaneously, the token guides the module to place greater emphasis on establishing robust cross-modal appearance representations. This, in turn, enhances the module's ability to perceive target appearance features during a modality switch. This interaction effectively associates the appearance feature spaces of the template and the search region, thereby mitigating the impact of appearance change caused by modality discrepancies.

\begin{figure}[htbp]
\centering
\includegraphics[width=\linewidth]{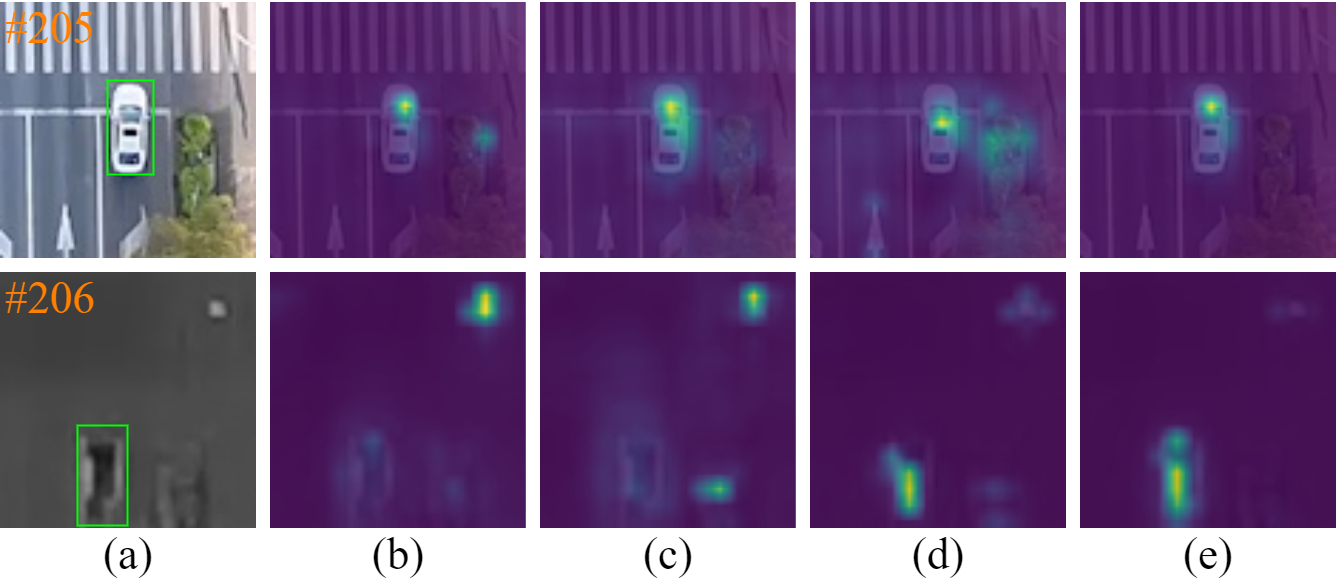}
\caption{Visualization of attention. (a) Search region with the ground truth. (b) Our baseline. (c) Our method with MSARM. (d) Our method with SSARM. (e) Our method with both modules.}
\label{fig:attention}
\end{figure}

As shown in Fig.~\ref{fig:attention} (c), even when the modality state remains unchanged, the modality state token still contributes positively to the module's feature extraction process. By continuously refining the feature representations of both the template and the search region, it enhances the robustness and discriminative power of the extracted features, ultimately improving the overall performance of the module.

In the training stage, the modality state token feature \(\mathbf{F}_{ms}\) is processed through a 3-layer MLP followed by a sigmoid function to estimate the current state.
To make the module possess the ability that module the appearance correlation, we use a loss to supervise the process.
The predicted modality state is compared with the ground truth, which is derived from the consistency between the modalities of the template frame and the search frame. 
Specifically, the ground truth indicates a consistent state when the modalities of the two frames are identical and an inconsistent state otherwise.
Upon completion of training, the modality state token acquires the capability to perceive and adapt to significant appearance change.   

\subsection{Spatial State Aware Representation Module}

To address the challenge of spatial shift, we propose the Spatial State Aware Representation Module (SSARM). SSARM introduces a spatial shift token and the reference region $I_{ref}$, which is cropped from the last frame and has the same position and size as the search region, to perceive the spatial state and model the cross-frame spatial correlation. Through the Transformer encoder, the spatial shift token establishes profound associations with both regions. 
During this process, the features of the search region $I_{z}$ and the reference region $I_{ref}$ are extracted. Simultaneously, via the attention operation, the mutual enhancement of the target-related feature representations of the two regions is accomplished.

Under the guidance of the spatial state token $\mathbf{F}_{ss}$, the module endeavors to associate the features of the search region $I_{z}$ with those of the reference region $I_{ref}$ to perceive the target movement within a frame. Subsequently, this association drives the module to gradually establish the cross-frame spatial correlation. The spatial state token functions as a signal, guiding the module to devote more attention to the association process. With the aid of this token, the module incorporates spatial perception and makes the necessary adjustments to adapt to the changes in the target position.
During the training stage, the spatial state token feature \(\mathbf{F}_{ss}\), obtained through multiple Transformer encoder layers, is subsequently passed into a shared 3-layer MLP followed by a sigmoid function to estimate the resulting spatial shift state.
Note that during testing, a memory bank is employed to store the previous frame, which is updated at each iteration of the module. The reference region is then cropped from the frame stored in the memory bank based on the tracking result from the previous step.

As illustrated in Fig.~\ref{fig:attention}, the existence of significant appearance change and sudden spatial shift presents a formidable challenge to our baseline model. During the process of modality switch, the model struggles to maintain a consistent focus on the target, with a consequent loss of attention.
After the modality switch, the model erroneously directs its attention to a white artifact that resembles the color of the previous target. This misdirection is notable, considering that the current target is predominantly black. Such a deviation from the intended target highlights the limitations of the baseline model in dealing with appearance change.
Owing to the lack of spatial shift perception, the model with MSARM fails to effectively focus on the target.
In contrast, the model with SSARM gains the ability to perceive spatial shift. This development facilitates a more focused attention on the target, as the model can now better account for changes in the target's position. Nevertheless, the model still shows some tendency to be distracted by the white artifact.
Ultimately, when both modules are used in combination, the model demonstrates an enhanced propensity to focus on the target. The result manifests the effectiveness of the MSARM. The combined effect of both modules reduces the model's susceptibility to distractions, leading to a more stable and accurate tracking performance.

\subsection{Loss Function}
Similarly to OSTrack~\cite{ye2022joint}, we utilize the weighted focal loss $L_{cls}$ for classification and the $L_{1}$ loss and the generalized IoU loss $L_{giou}$ for bounding box regression.
Furthermore, our model includes an auxiliary output that predicts spatial shift $(\mathbf{p}_{\Delta x},\mathbf{p}_{\Delta y})$. 
Specifically, the new output head is implemented as a fully convolutional network (FCN), which comprises four stacked Conv-BN-ReLU layers. Each convolutional layer utilizes a \(3 \times 3\) kernel to capture spatial features, followed by a \(2 \times 2\) max-pooling operation to reduce computational complexity.

The ground truth of the spatial shift is expressed as $(\mathbf{gt}_{\Delta x},\mathbf{gt}_{\Delta y})$.
This spatial shift quantifies the offset of the target center, defined by the coordinate pairs $(x_{c}^{t - 1},y_{c}^{t - 1})$ and $(x_{c}^{t},y_{c}^{t})$ for the previous and current frames respectively, where $x_{c}$ and $y_{c}$ are the $x$ and $y$ coordinates of the target center and the superscripts denote the frame number. The spatial shift is calculated as follows:
\begin{equation}
    \begin{aligned}
        (\mathbf{gt}_{\Delta x}, \mathbf{gt}_{\Delta y}) &= (x_{c}^{t},
        y_{c}^{t})-(x_{c}^{t-1},y_{c}^{t-1}).
    \end{aligned}
    \label{eq:spatial_shift}
\end{equation}
To measure the difference between the predicted and ground truth spatial shift, the Mean Absolute Error (MAE) $L_{ssp}$ is employed, and its calculation formula is given by:
\begin{equation}
    \begin{aligned}
        L_{ssp} &= MAE((\mathbf{p}_{\Delta x},\mathbf{p}_{\Delta y}),(\mathbf{gt}_{\Delta x}, \mathbf{gt}_{\Delta y})).
    \end{aligned}
\end{equation}
\textcolor{red}{During training, the predicted spatial shift $(\mathbf{p}_{\Delta x}, \mathbf{p}_{\Delta y})$ serves as an auxiliary supervision signal for SSARM. It is used to guide the spatial-state token $\mathbf{F}_{ss}$ to learn the displacement relationship between the reference-frame region and the current search region. Notably, this predicted shift is not used to explicitly adjust the final bounding box or relocate the search region during inference. The final tracking result is still produced by the original prediction head. Therefore, the spatial shift prediction improves tracking performance indirectly by enhancing the spatial-aware representation learned by $\mathbf{F}_{ss}$, rather than acting as a post-processing correction.}

Since both the modality state and the spatial state have two classes for classification, the BCEWithLogitsLoss is adopted for $L_{ms}$ and $L_{ss}$, respectively. 
The overall loss function is:
\begin{equation}
    \begin{aligned}
        L_{total} = &L_{cls} + \lambda _{giou}L_{giou}+\lambda _{L1}L_{1} +L_{ssp} + \\
         & L_{ms} + L_{ss}
    \end{aligned}
\end{equation}
where $\lambda _{giou}$ = 2 and $\lambda _{L1}$ = 5 are the regularization parameters in our experiments as in OSTrack~\cite{ye2022joint}.

\section{Benchmark Dataset}

\subsection{Data Collection}
This dataset is constructed using professional UAVs, specifically the DJI Matrice 300 RTK equipped with a Zenmuse H20T thermal imaging camera and the DJI Mavic 3E.
The dataset is designed to reflect the operational characteristics of UAVs in real-world scenarios, particularly their high mobility and wide field of view. Data collection is conducted at altitudes ranging from 5 meters to 500 meters, ensuring coverage of both fine-grained details and broader spatial contexts.
To ensure temporal and environmental diversity, the data capture process spans multiple seasons and includes both daytime and nighttime operations. Geographic coverage is extensive, encompassing a variety of urban and natural landscapes, including schools, parks, rivers, zoos, and diverse street types.

\begin{figure}[htbp]
\centering
\includegraphics[width=\linewidth]{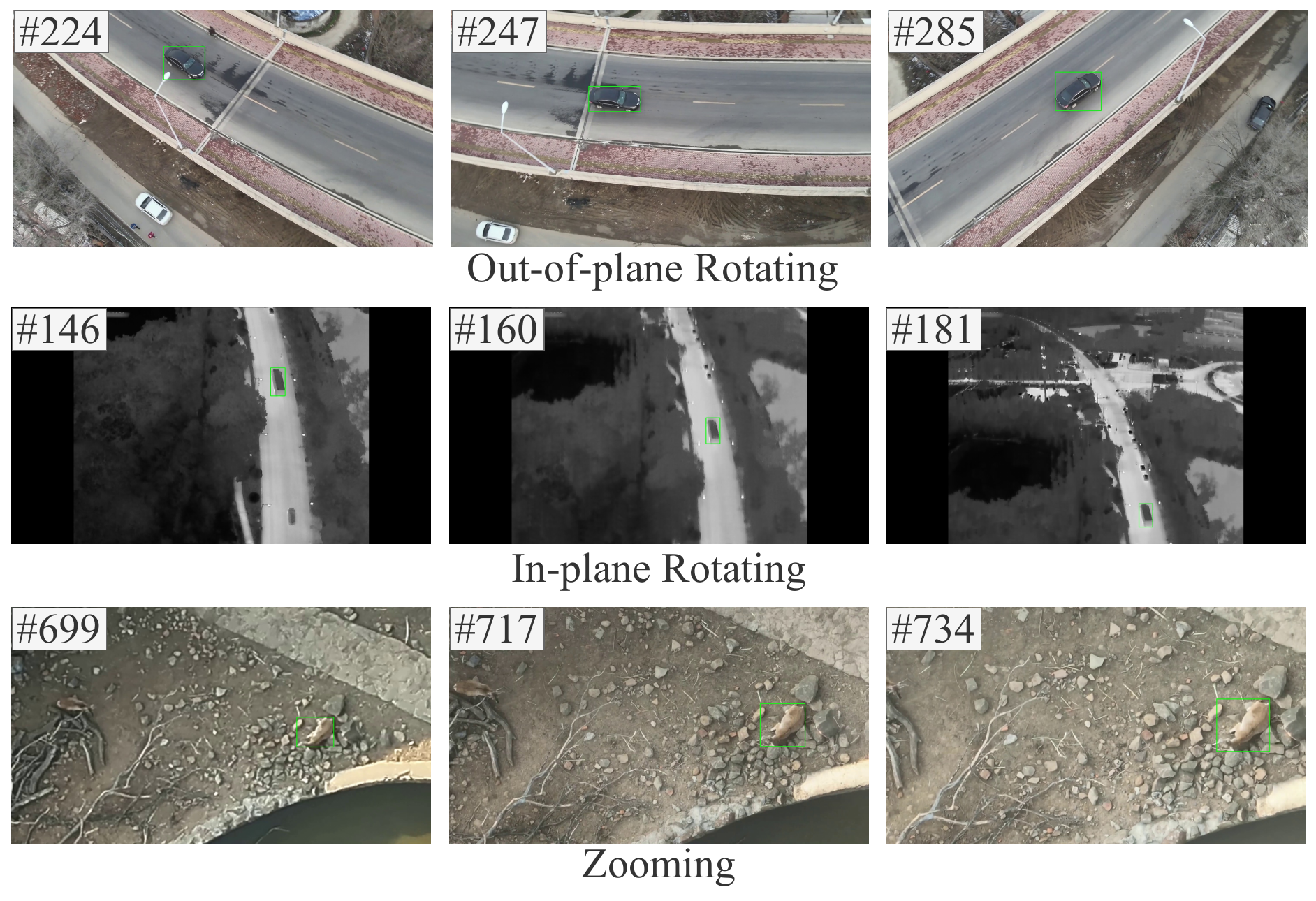}
\caption{Sample frames from the CM-UOT dataset. Each row shows a type of UAV movement.}
\label{fig:movement}
\end{figure}

Additionally, dynamic scenarios such as sports activities in playgrounds are included to capture fast-moving objects and complex interactions.
To further enhance the dataset's applicability, data is collected under extreme weather conditions, including high winds, snowfall, and fog. 
% These conditions challenge the operational limits of UAVs and provide valuable data for developing algorithms capable of functioning reliably in adverse environments.
In alignment with typical UAV operational patterns, the data capture process incorporates a range of maneuvers, including linear movements (e.g., forward, backward, and lateral flights), rotational movements to adjust viewing angles, and zooming actions to focus on specific targets. 
These maneuvers are systematically executed to simulate real-world UAV operations and to ensure the dataset's relevance for practical applications. Some examples are shown in Fig.~\ref{fig:movement}.

\begin{figure}[htbp]
\centering
\includegraphics[width=\linewidth]{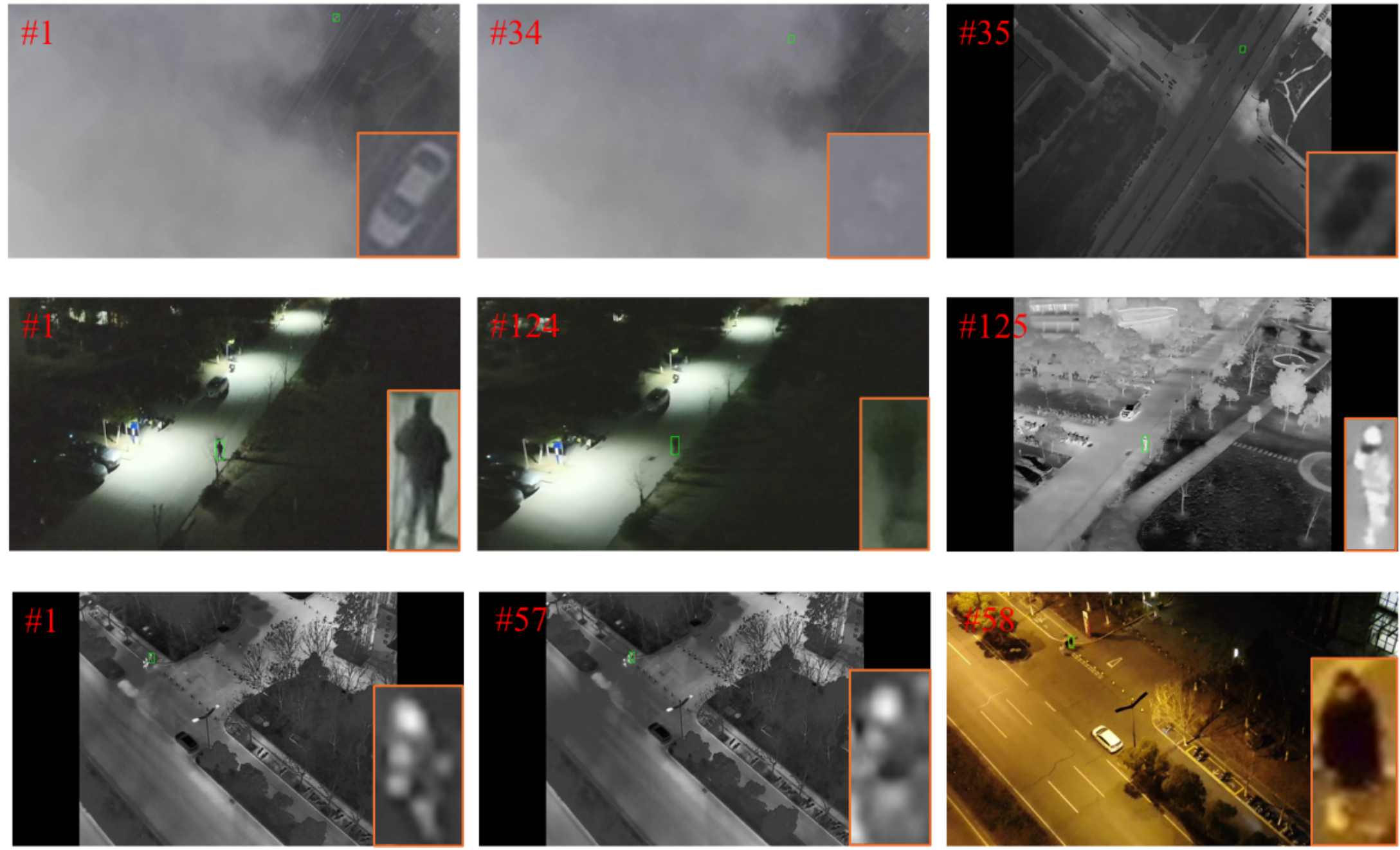}
\caption{Sample frames from the CM-UOT dataset. The red box at the bottom right is a local magnification for clearer visualization of the target appearance.}
\label{fig:switch}
\end{figure}

We thoroughly consider the imaging quality of both modalities. Fig.~\ref{fig:switch} shows some typical examples of the CM-UOT dataset.
Finally, we carefully selected 1,079 sequences to confront the challenges encountered in real life, with a total of over 673K frames and an average video length of more than 621 frames. The images are of high quality, stored in JPG format with a resolution of 1920$\times$1080 and sampled at 30 fps. 

\subsection{Annotation}
In our dataset, we provide annotations for bounding boxes with absent labels specifically referring to LaSOT~\cite{fan2019lasot,Fan2021}. Thanks to the open-source nature of CVAT, a web-based image annotation tool originally developed by Intel and now maintained by OpenCV, we easily deploy a server to support our team's annotation work. The tool allows annotators to annotate bounding boxes and absent labels, including those that are out-of-view or fully occluded, all through an intuitive interface easily accessible via any web browser. 
To ensure the accuracy of annotation, we adopted a two-phase annotation strategy.
In the first phase, annotators precisely annotate the bounding box of the target and mark the absent state for cases where the target is fully occluded or out of view.
In the second phase, checkers carry out a comprehensive inspection to minimize the likelihood of incorrect or inaccurate labels.

\subsection{Attributes}
\begin{figure}[htbp]
\centering
\includegraphics[width=\linewidth]{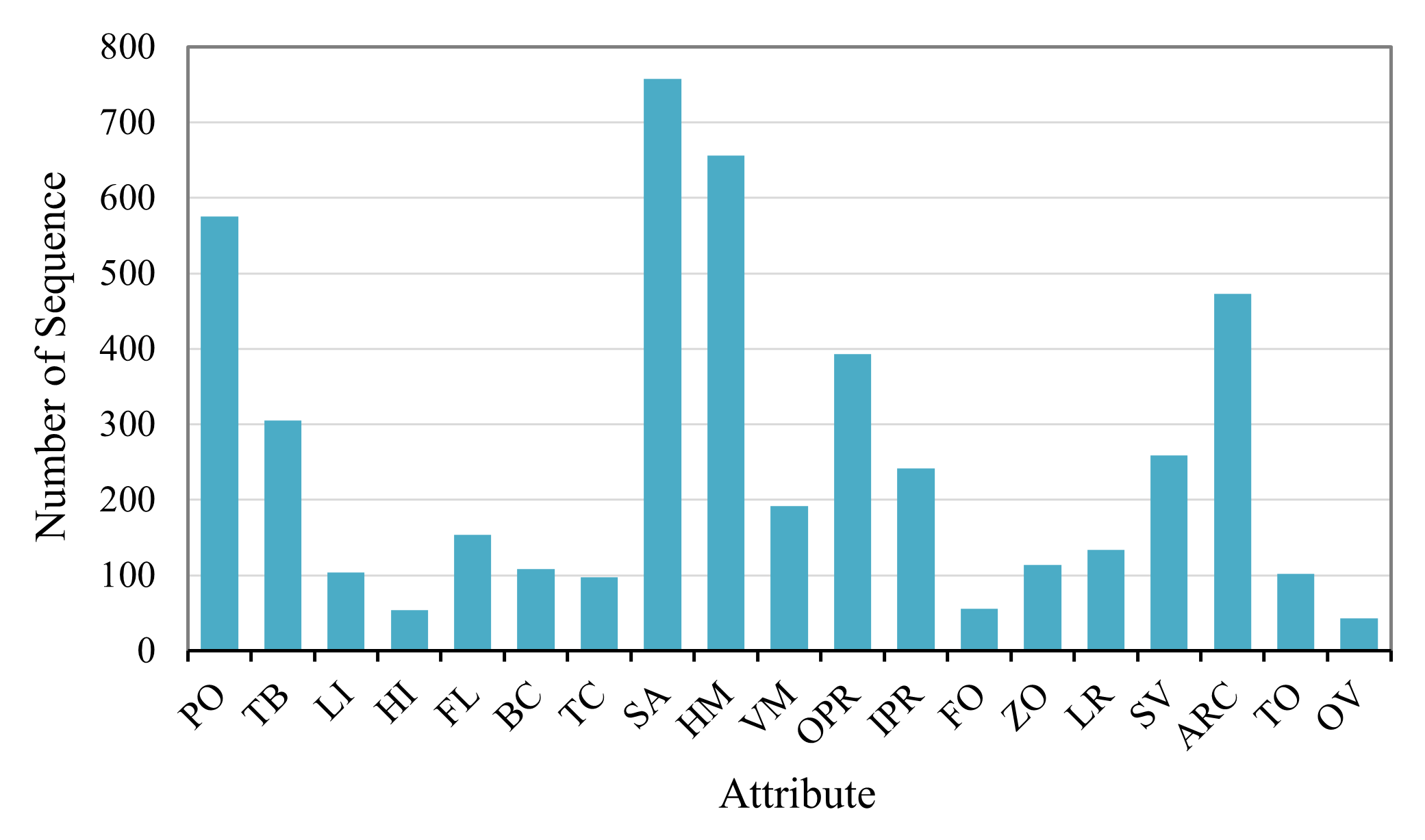}
\caption{Distribution of attributes on CM-UOT.}
\label{fig:attribute}
\end{figure}

Existing multi-modal tracking datasets incorporate two modalities in each frame; in contrast, our dataset, similar to the CMOTB dataset, features only one modality per frame, with potential modality switch occurring throughout the sequence. 
We thoughtfully incorporate numerous drone characteristics, making our dataset more challenging for cross-modal UAV object tracking. Therefore, we introduce 5 drone-specific attributes and 1 additional attribute for foggy conditions, and also annotate 13 common attributes, as shown in Fig.~\ref{fig:attribute}. To avoid ambiguity in defining certain attributes in the context of UAVs, such as rapid motion (RM), viewpoint change (VC), camera motion (CM), these factors are excluded from consideration. 

Challenges are summarized as 19 attributes, including partial occlusion (PO), target blur (TB), low illumination (LI), high illumination (HI), frame lost (FL), background clustering (BC), thermal crossover (TC), similar appearance (SA), horizontal movement (HM), vertical movement (VM),  out-of-plane rotation (OPR), in-plane rotation (IPR)~\cite{li2017visual}, fog (FO), zoom (ZO), low resolution (LR), scale variation (SV), aspect ratio change (ARC), total occlusion (TO) and out-of-view (OV). The description of attribute is summarized in Table~\ref{tab:attri}.

\begin{table}[htbp]
\caption{\textcolor{red}{Descriptions of attributes in the CM-UOT dataset.}}\label{tab:attri}
\begin{tabular}{@{}lp{0.85\linewidth}@{}}
\toprule
 & Definition \\
\midrule

PO & Partial Occlusion - the target is partially occluded. \\
TB & Target Blur - the target object motion results in the blur image information. \\
LI & Low Illumination - the illumination in the target region is low. \\
HI & High Illumination - the illumination in the target is too strong to identify the target. \\
FL & Frame Lost - some of thermal frames are lost. \\
BC & Background Clutter - the background information which includes the target object is messy. \\
TC & Thermal Crossover - the target has similar temperature with other objects or background surroundings. \\
SA & Similar Appearance - there are objects of similar appearance near the target. \\
\textbf{HM} & Horizontal Movement - the UAV executes horizontal movements. \\
\textbf{VM} & Vertical Movement - the UAV executes vertical movements. \\
\textbf{OPR} & Out-of-plane Rotation - the UAV or its camera yaw rotates. \\
\textbf{IPR} & In-plane Rotation - the UAV's camera pitch rotates. \\ 
\textbf{FO} & Fog - the weather is foggy. \\
\textbf{ZO} & Zoom - the UAV's camera zooms in or out. \\
LR & Low Resolution - bounding box area less than 400. \\ 
SV & Scale Variation - ratio of current to initial bounding box outside $\tau \in [0.5, 2]$. \\
ARC & Aspect Ratio Change - significant change in bounding box aspect ratio outside $[0.5, 2]$. \\
TO & Total Occlusion - the target is totally occluded. \\
OV & Out-of-View - the target leaves the camera field of view. \\
\bottomrule
\end{tabular}
\end{table}

\begin{figure}[htbp]
\centering
\includegraphics[width=\linewidth]{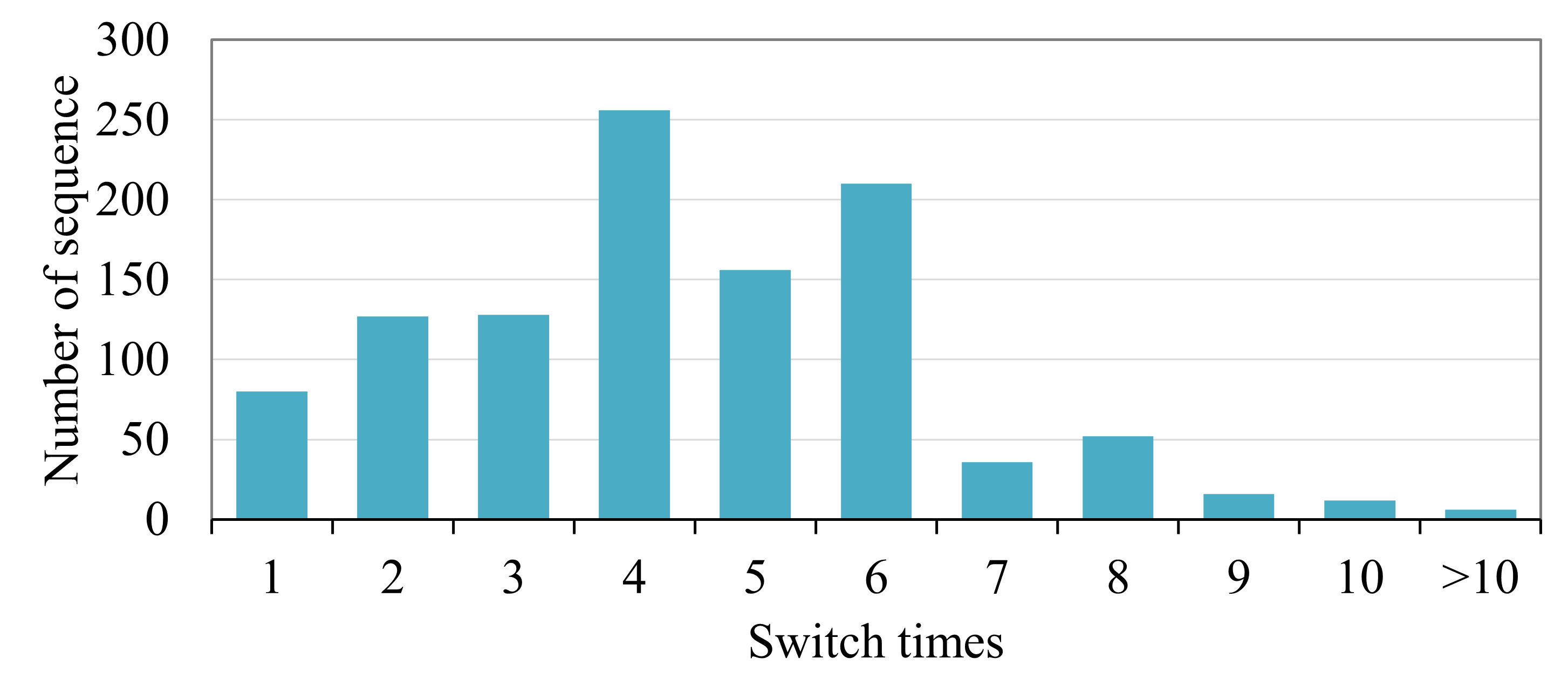}
\caption{Distribution of the number of modality switch on CM-UOT.}
\label{fig:switchtime}
\end{figure}

The number of modality switches within a sequence stands as a crucial factor influencing the performance of trackers, reflecting the frequency of significant appearance alterations and abrupt spatial displacements.
For the CMOTB dataset, the average number of modality switches per sequence is 1.6, with a maximum of four such switches occurring in any given sequence. In contrast, the CM-UOT dataset demonstrates a considerably higher average, with 4.4 modality switches per sequence. Moreover, the maximum number of modality switches observed in a single sequence within the CM-UOT dataset is 21.
Consequently, during the data creation process, we meticulously record the number of modality switches and present the distribution of these switch counts in Fig. \ref{fig:switchtime}.

\subsection{Statistics}
\begin{figure}[htbp]
\centering
\includegraphics[width=\linewidth]{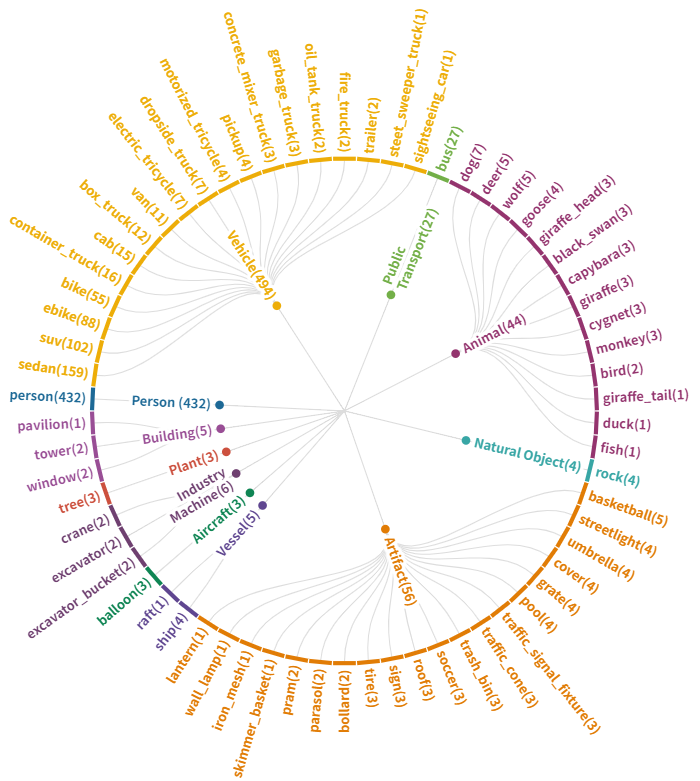}
\caption{Distribution of object categories in the CM-UOT benchmark dataset. }
\label{fig:class}
\end{figure}

To facilitate model training and evaluation, we divided the dataset into training and testing sets in an 80:20 ratio, with 863 sequences in the training set and 216 sequences in the testing set. 
To enhance our dataset, we draw inspiration from the category structure of WebUAV~\cite{zhang2022webuav}, ultimately selecting 66 categories, making a few adjustments based on WordNet~\cite{miller1995wordnet}. 
In line with that dataset, we organize our categories into 11 superclasses (e.g., person, vehicle, vessel, animal, plant, artifact, and natural object), as illustrated in Fig. \ref{fig:class}.
Notably, we have expanded the person superclass by incorporating both the target class (person) and a specialized motion class. 
The motion class encompasses 12 distinct human activity categories: walking, running, crossing the road, playing basketball, engaging in play activities, working, playing soccer, climbing stairs, biking, playing tennis, throwing, and getting off.

\subsection{Aligned CM-UOT}
In our cross-modal tracking task, modality switch not only induces significant appearance change but also results in sudden spatial shift in target position.
Such shift is attributed to the inherent positional discrepancies between the two sensors, which are affected in different ways by independently adjustable focal lengths, changes in flight heights, and camera orientation adjustments.
The two primary challenges are closely intertwined, making it difficult to conduct targeted research on each individually.
To help researchers systematically tackle these challenges, we provide an aligned version of CM-UOT that eliminate the spatial shift caused by modality switch.

\begin{table*}[htbp]
    \caption{
        Statistics comparison among existing cross-modal object tracking datasets and multi-modal UAV object tracking datasets. Modality: (R)GB, (T)hermal-IR and (N)ear-IR; Task could be either (C)ross-modal and (M)ulti-modal object tracking;    Platform could be either (A)erial or (G)round-based.
    }\label{tab:benchmark}
    \centering
    \setlength{\tabcolsep}{4pt}
    \resizebox{\textwidth}{!}{%
    \begin{tabular}{l|ccccccccccc}
            \toprule
            Benchmark & Modality & Task & Platform  & Video & Class & Attributes   & \makecell{Motion\\Class} & Resolution & Altitude   & \makecell{Absent\\Label} & \makecell{Switch\\Times} \\
            \midrule
            VTUAV & R, T & M & A & 500  & 13 & 13 & $\times$  & 1920$\times$1080 & 5-20 m & $\times$ & - \\
            CMOTB  & R, N & C & G & 1000  & 61 & 18 & $\times$ & 1280$\times$720 & -   & $\times$  & 1.6 \\
            \midrule
            CM-UOT & R, T & C & A & 1079  & 66 & 19 & $\checkmark$  & 1920$\times$1080 & 5-500 m  & $\checkmark$ & 4.4 \\
            \bottomrule       
    \end{tabular}
    }
\end{table*}

\subsection{Comparison with Existing Datasets}
Cross-modal UAV object tracking presents unique challenges and considerations that distinguish it from traditional tracking scenarios. This can be observed in several aspects, with existing benchmarks and datasets having limitations when applied to cross-modal UAV object tracking, and our work addressing these gaps through a novel dataset. The statistics comparison among them as shown in Table~\ref{tab:benchmark}.

CMOTB~\cite{li2022cross,liu2024cross} is a pioneering benchmark for cross-modal object tracking, utilizing surveillance cameras that switch between RGB and near-infrared (NIR) imaging based on light intensity. 
Although it has advanced cross-modal tracking research, its effectiveness in evaluating UAV cross-modal tracking algorithms is limited due to several key differences.
First, UAVs typically use TIR sensors, which detect heat signatures, unlike NIR sensors that rely on reflected light. This results in distinct modality characteristics, as seen in their differing transmission behaviors through glass (NIR penetrates, TIR does not).
Second, the scenarios where UAVs operate have certain unique attributes, including the presence of tiny objects, intense motion, rotation, and zooming.
Finally, the RGB and TIR sensors on UAVs typically employ separate cameras, which inevitably gives rise to spatial shift. Such shift is further aggravated by the dynamic movement of the UAV, rendering such deviations highly challenging to predict and rectify. 

The task of multi-modal (RGBT) UAV object tracking is also relevant to our work.
VTUAV ~\cite{zhang2022visible}, a RGBT UAV object tracking dataset, has been released. It contains 383 short-term sequences and 117 long-term sequences. 
There are several differences between VTUAV and our dataset, which are elaborated as follows:
First, considering the constraints imposed by UAVs in our specific task, our dataset solely provides one modality's data at each frame time.
Second, due to the occurrence of modality switch within a single frame in our task, our dataset adopts a dense annotation strategy.
Finally, our dataset comprehensively accounts for UAV characteristics, including flight altitude and motion patterns.

\section{Experiment}
In this section, we adopt three evaluation protocols to comprehensively assess the proposed SARLA method for cross-modal UAV object tracking.
Specifically, we first determine the evaluation metrics (OPE protocol, PR, NPR, SR) to ensure consistent performance assessment. 
We then conduct a series of experiments on the self-constructed CM-UOT dataset, which include comprehensive performance evaluation of 20 superior tracking methods, comprehensive validation of 8 representative trackers retrained on this dataset, attribute-based performance evaluation, quantitative analysis of the impact of spatial shift magnitude, and assessment of the impact of modality switch by splitting sequences. 
Additionally, we perform ablation studies to verify the effectiveness of key components (MSARM, SSARM, SSPL) in SARLA, present qualitative comparisons on representative sequences with significant appearance change and spatial shift, and further validate SARLA’s generalization on the public CMOTB dataset.
\subsection{Evaluation Metrics}
In our experiments, we adopt the one-pass evaluation (OPE) protocol to ensure consistency in tracker performance assessment, and use three key metrics widely recognized in tracking~\cite{muller2018trackingnet}: precision rate (PR), normalized precision rate (NPR), and success rate (SR).

\begin{itemize}
\item Precision Rate (PR): Measures the percentage of frames where the predicted position is within a 20-pixel threshold of the ground truth center.
\item Normalized Precision Rate (NPR): Adjusts PR by bounding box size to fairly evaluate trackers across objects of different scales.
\item Success Rate (SR): Evaluates tracking via frame-by-frame IoU between predictions and ground truth. The final score is the area under the IoU threshold curve (AUC), providing a scale-independent accuracy measure.
\end{itemize}

\begin{figure*}[htbp]
\centering
\includegraphics[width=0.32\textwidth]{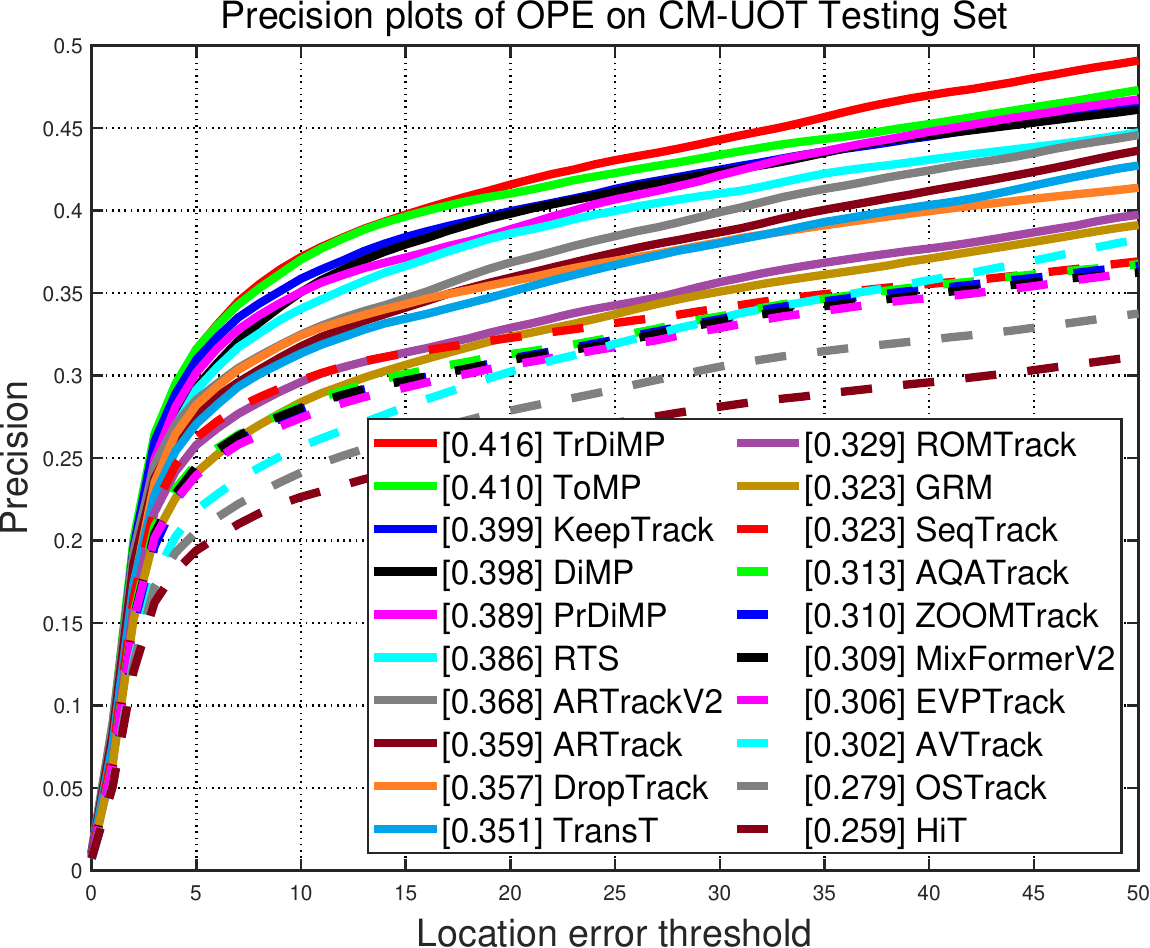}
\hfill
\includegraphics[width=0.32\textwidth]{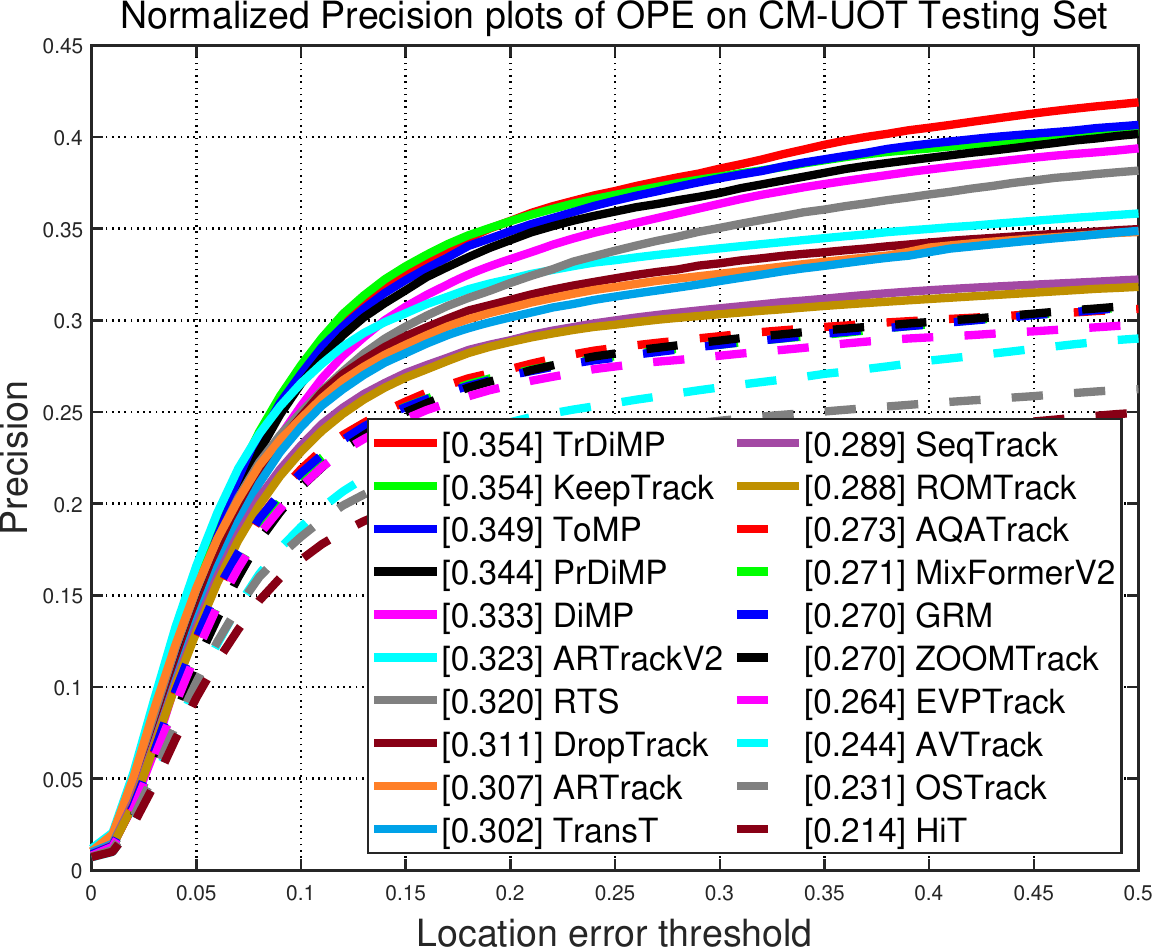}
\hfill
\includegraphics[width=0.32\textwidth]{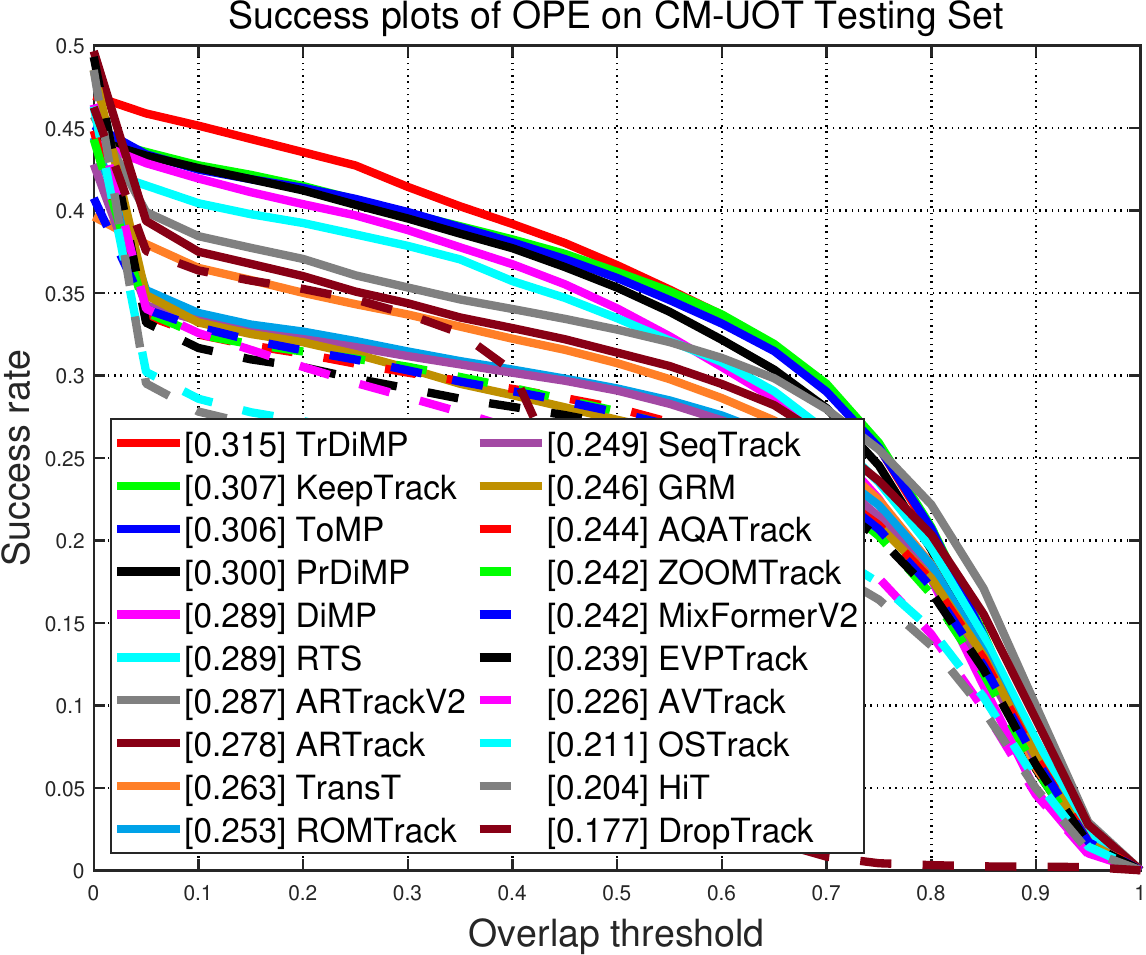}
\caption{Evaluation result on entire CM-UOT using precision, normalized precision and success plots, where the representative scores are presented in the legend.}
\label{fig:evaluation}
\end{figure*}

\subsection{Implementation Details}
We employ the base version of ViT-B in the feature extraction network. We conduct an end-to-end training process. In terms of input data, we take a template with $128 \times 128$, a reference region with $256 \times 256$, and a search region with $256 \times 256$, where the template-search pairs are randomly sampled and the search frames are adjacent to the reference frames. We train the model with AdamW optimizer, set the weight decay to $10^{-4}$, the initial learning rate for the backbone to $4 \times 10^{-5}$, and other parameters to $4 \times 10^{-4}$, respectively. The total training epochs are set to 50 with 60K image pairs per epoch, and we decrease the learning rate by a factor of 10 after 40 epochs. The model is implemented on the PyTorch \cite{NEURIPS2019_bdbca288} platform and runs on a single Nvidia RTX 3090 GPU with 24G memory.

\subsection{Evaluation on CM-UOT}
\noindent
\textbf{Overall Performance}. We evaluated 20 most advanced and representative trackers on our benchmark. These trackers cover mainstream tracking algorithms from 2019 to 2024, and they are 
DiMP~\cite{bhat2019learning}, 
PrDiMP\allowbreak~\cite{danelljan2020probabilistic}, 
TransT~\cite{chen2021transformer}, 
TrDiMP\allowbreak~\cite{wang2021transformer}, 
KeepTrack~\cite{mayer2021learning}, 
RTS~\cite{paul2022robust}, 
ToMP~\cite{mayer2022transforming}, 
OSTrack~\cite{ye2022joint}, 
HiT~\cite{kang2023exploring}, 
ZoomTrack~\cite{kou2023zoomtrack}, 
SeqTrack~\cite{chen2023seqtrack}, 
MixFormerV2\allowbreak~\cite{cui2023mixformerv}, 
ROMTrack~\cite{cai2023robust}, 
DropTrack~\cite{wu2023dropmae}, 
GRM~\cite{gao2023generalized}, 
ARTrack~\cite{wei2023autoregressive}, 
ARTrackV2\allowbreak~\cite{bai2024artrackv2}, 
AQATrack~\cite{xie2024autoregressive}, 
AVTrack~\cite{li2024learning},
EVPTrack~\cite{shi2024explicit}.
Note that all algorithms are evaluated on our test set using the model provided by the authors.

We first present the evaluation results on the CM-UOT dataset, where 216 sequences are used as testing set for large-scale evaluations. We summarize the overall evaluation results of these trackers with the precision, normalized precision, and success plots, as shown in Fig.~\ref{fig:evaluation}.
From the results, we can see that TrDiMP achieves the best precision score of 41.6\%, normalized precision score of 35.4\%, and success score of 31.5\%. However, the performance is limited in our task due to the existence of large heterogeneous gap between RGB and TIR modalities and abrupt spatial shift. They remain substantial room for improvement.

\begin{figure*}[tbp!]
\centering
\includegraphics[width=0.9\linewidth]{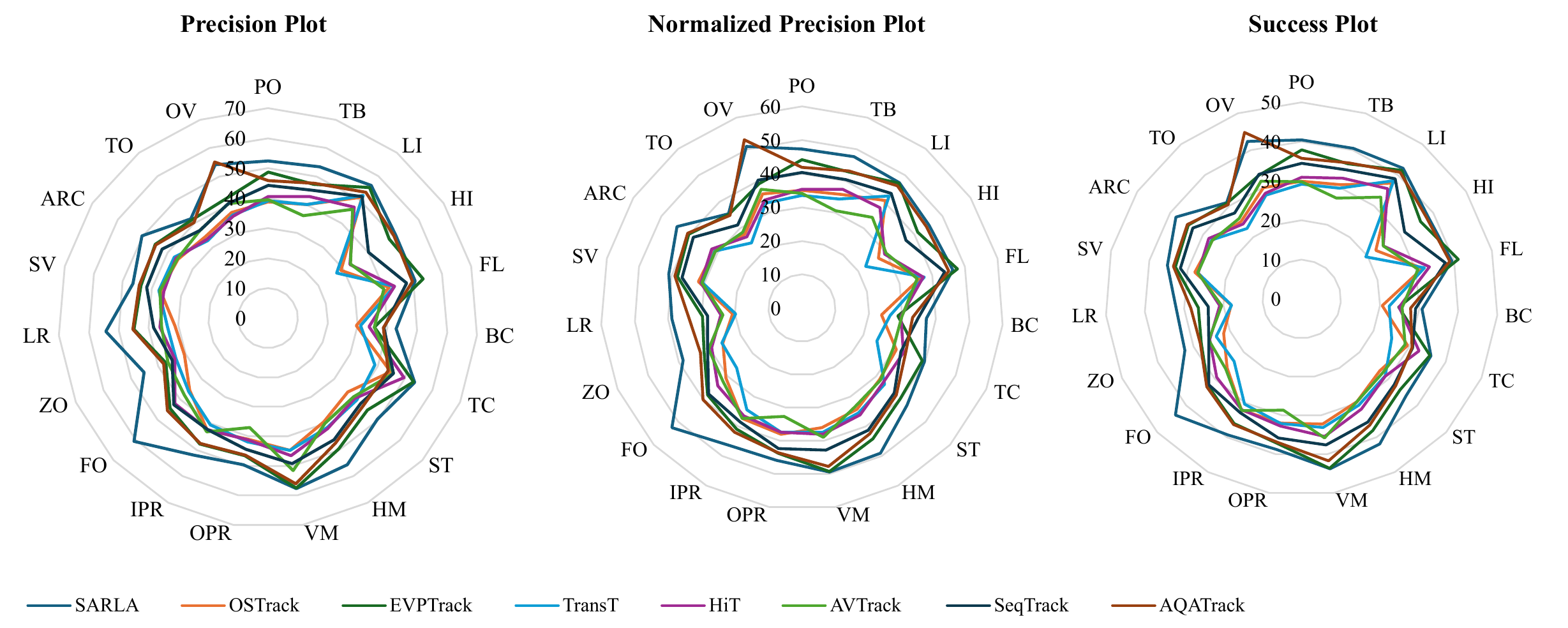}
\caption{\textcolor{red}{Complete attribute-based comparison on all 19 annotated attributes of the CM-UOT dataset, reporting the PR, NPR and SR metrics.}}
\label{fig:all_attr}
\end{figure*}

\noindent
\textbf{Training Dataset Validation}. We select 10 representative trackers including 
TransT~\cite{chen2021transformer},
OSTrack~\cite{ye2022joint},
HiT~\cite{kang2023exploring},
SeqTrack~\cite{chen2023seqtrack},
GRM~\cite{gao2023generalized},vijp
EVPTrack~\cite{shi2024explicit},
AVTrack~\cite{li2024learning},
AQATrack~\cite{xie2024autoregressive},
SUTrack~\cite{chen2025sutrack}, and SpikeTrack~\cite{zhang2026spiketrack}to demonstrate the effectiveness of our training dataset in the training of deep models.Each model is consistently trained for 50 epochs. 
The results are shown in Table~\ref{tab:retrain}, which shows that all the re-trained deep trackers achieve obvious improvements and verify the significant necessity of proposing this dataset for the study of cross-modal object tracking. Specially, SARLA achieves the top performance with 50.6\% PR, 44.8\% NPR and 38.0\% SR.

\noindent
\textbf{Attribute-based Performance}.
\textcolor{red}{As shown in Fig.~\ref{fig:attribute}, the attribute distribution of CM-UOT is naturally imbalanced. Frequent challenges, such as similar appearance, horizontal movement, partial occlusion, and aspect ratio change, contain a relatively large number of sequences, whereas rare challenges, such as high illumination, fog, total occlusion, and out-of-view, contain much fewer samples. Such imbalance may affect the statistical stability of attribute-wise evaluation, since the results on rare attributes are more sensitive to individual sequences and may exhibit larger fluctuations. Nevertheless, the overall tracking accuracy is not directly biased by this imbalance, because all trackers are evaluated on the same test set under the same attribute distribution. Therefore, the attribute-wise results for rare categories are interpreted as supplementary evidence and are analyzed together with the overall performance and other attribute groups. As shown in Fig.~\ref{fig:all_attr}, SARLA achieves competitive performance across most attributes, indicating that its advantage is not limited to several high-frequency attributes but generalizes to diverse UAV-specific and general tracking challenges.}

\begin{figure*}[tbp!]
\centering
\includegraphics[width=0.9\linewidth]{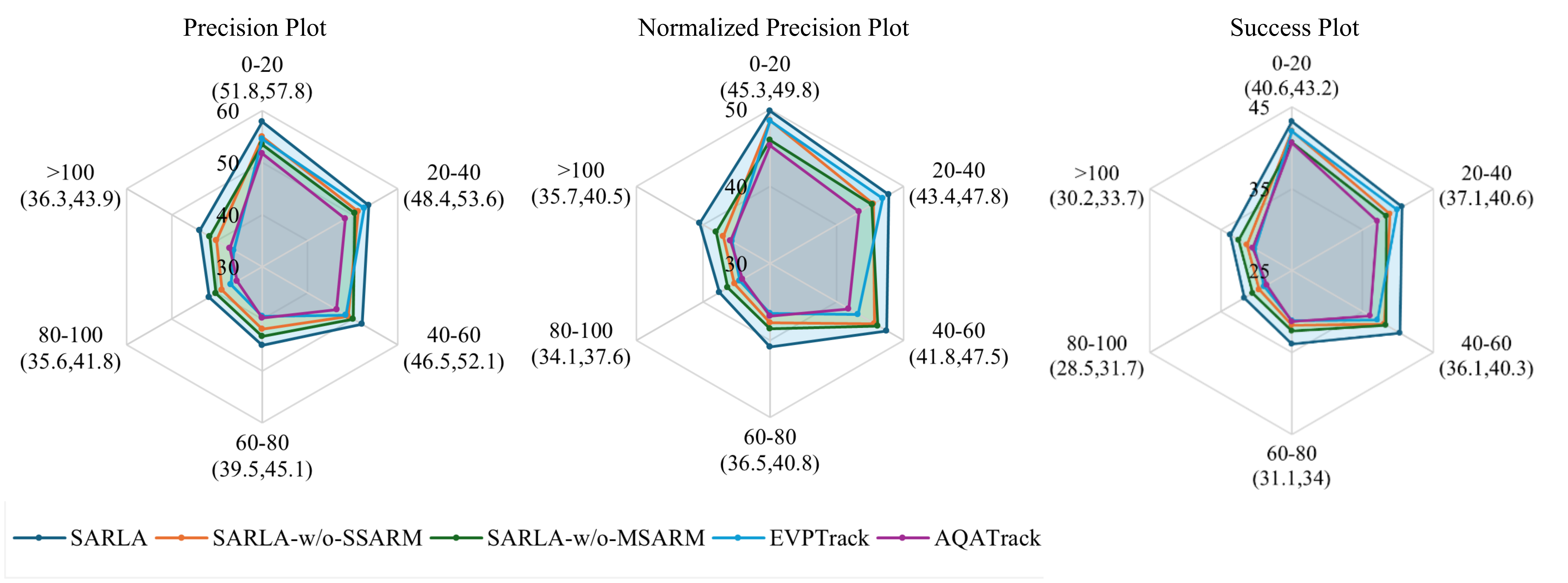}
\caption{Comparison based on spatial shift on CM-UOT testing set.}
\label{fig:offset}
\end{figure*}

\noindent
\textbf{Impact of Spatial Shift}.
The magnitude of spatial shift induced by modality switch is a critical factor governing the performance of visual trackers.
To quantitatively evaluate the impact of spatial shift magnitude on tracker performance, we stratified the dataset into six groups according to the target center offset, with the ranges set as 0–20, 20–40, 40–60, 60–80, 80–100, and $>$  100 pixels, respectively.
Corresponding performance metrics for each group are presented in Fig.~\ref{fig:offset}, where only the results of two representative competing trackers are displayed for clarity.

\begin{table*}[htbp]
\caption{The PR, NPR, SR scores (\%) of various trackers on different parts. The parts include 3 types: (\textbf{Before}) the first modality switch, (\textbf{After}) the first switch and \textbf{All}.}\label{tab:split}
\centering
\resizebox{0.8\textwidth}{!}{%
\begin{tabular}{l|c|lll|lll|lll}
    \toprule
        \multirow{2}{*}{Method} & \multirow{2}{*}{Pub. Info.} & ~ & PR & ~ & ~ & NPR & ~ & ~ & SR & ~ \\ 
        & & \textbf{Before} & \textbf{After} & \textbf{All} & \textbf{Before} & \textbf{After} & \textbf{All} & \textbf{Before} & \textbf{After} & \textbf{All} \\ 
        \midrule
        TransT & CVPR21 & 90.3 & 31.0 & 39.5 & 82.9 & 25.6 & 33.6 & 68.6 & 22.7 & 29.2 \\ 
        OSTrack & ECCV22 & 89.1 & 27.7 & 37.0 & 84.8 & 23.8 & 32.7 & 70.2 & 20.7 & 28.0 \\ 
        HiT & NeurIPS23 & 87.4 & 31.7 & 40.0 & 81.1 & 26.7 & 34.3 & 68.2 & 23.8 & 30.1 \\ 
        SeqTrack & CVPR23 & 90.9 & 34.7 & 43.3 & 87.1 & 30.9 & 39.2 & 72.8 & 26.6 & 33.5 \\ 
        GRM & CVPR23 & 89.1 & 29.4 & 38.5 & 84.4 & 24.8 & 33.5 & 70.1 & 21.7 & 28.8 \\ 
        EVPTrack & AAAI24 & \textbf{91.7} & 38.3 & 46.5 & \textbf{87.6} & 33.6 & 41.7 & \textbf{72.8} & 29.1 & 35.7 \\  
        AVTrack & ICML24 & 89.0 & 29.0 & 38.0 & 82.0 & 23.7 & 32.0 & 68.4 & 21.3 & 28.1 \\ 
        AQATrack & CVPR24 & 89.9 & 36.3 & 44.5 & 85.3 & 32.3 & 40.1 & 71.1 & 27.8 & 34.2 \\ 
        \midrule
        SARLA & - & 91.1 & \textbf{43.3} & \textbf{50.6} & 86.5 & \textbf{37.5} & \textbf{44.8} & 71.7 & \textbf{32.2} & \textbf{38.0} \\ 
        \bottomrule
    \end{tabular}
}
\end{table*}

As observed from the figure, the performance of all trackers exhibits a consistent downward trend with increasing offset values.
This phenomenon quantitatively verifies that spatial shift magnitude exerts a significant influence on tracking performance.
Notably, our proposed method SARLA achieves markedly smaller performance degradation compared to other trackers across all groups.
Furthermore, distinct performance patterns are observed between the two variants:  SARLA-w/o-MSARM (with the MSARM module removed) outperforms SARLA-w/o-SSARM (with the SSARM module removed) in the first two groups, where the relatively small spatial shift render the modality gap the dominant challenge for tracking accuracy.
In contrast, SARLA-w/o-SSARM demonstrates superior performance in the remaining four groups, where large spatial shift emerge as the primary performance bottleneck.
In such scenarios, the spatial correlation modeling capability for which SSARM is specifically designed becomes far more critical than modality gap mitigation.

These variant-specific performance trade-offs directly highlight the complementary strengths of MSARM and SSARM: the former excels at addressing modality discrepancies in small-shift scenarios, while the latter specializes in handling spatial shift under large-offset conditions. 
Building on this synergy, these results further confirm that the integrated deployment of both modules enables SARLA to simultaneously tackle both core challenges.
Consequently, SARLA achieves the most robust and consistent performance across all six offset groups.
This result conclusively validates that our proposed framework is highly effective at addressing the challenges posed by large spatial shift in cross-modal tracking scenarios.

\noindent
\textbf{Impact of Modality Switch}.
Before the first modality switch occurs, the tracking process is typically free from significant appearance change and spatial shift induced by modality switch. Consequently, the moment at which the first modality switch takes place exerts a significant influence on tracking performance.
Analyzing result according to the proportion of the initial switching time to the total video duration is a reasonable way to prove that. However, since this ratio is continuous and the number of test sequences is limited, it is difficult to analyze them effectively. 

To clearly illustrate the impact of modality switch, we divide each sequence into two parts at the first modality switch timestamp and calculate metrics for each part independently (as opposed to conventional per-sequence metric computation). As shown in Table~\ref{tab:split}, all trackers experience significant performance degradation after the first switch, which reflects the substantial challenges induced by sudden appearance variations and spatial shift. Prior to the switch, EVPTrack leads the rankings, followed by SeqTrack and our method. Post-switch, our method attains top performance, surpassing the second-ranked EVPTrack by $5.0\%$/$4.1\%$/$3.1\%$ in PR/NPR/SR and the baseline OSTrack by $15.6\%$/$13.7\%$/$11.5\%$ in the same metrics. This confirms our method's effectiveness in addressing the aforementioned issues. Additionally, our method outperforms the baseline by $2\%$/$1.7\%$/$1.5\%$ in PR/NPR/SR even before the switch, verifying its positive impact on non-cross-modal tracking.

\begin{table}[t]
\centering
\caption{Comparison with global-search and re-detection trackers on CM-UOT.}
\label{tab:global_redetection}
\begin{tabular}{lcccc}
\hline
Method & Strategy & PR & NPR & SR \\
\hline
SPLT & Re-detection & 37.4 & 32.8 & 28.1 \\
GlobalTrack & Global search & 39.8 & 34.2 & 30.0 \\
ZoomTrack & Adaptive search & 47.1 & 41.6 & 35.2 \\
SARLA & State-aware learning & \textbf{50.6} & \textbf{44.8} & \textbf{38.0} \\
\hline
\end{tabular}
\end{table}

\textcolor{red}{\textbf{Comparison with Global-Search and Re-detection Trackers.}}
\textcolor{red}{In addition to conventional short-term trackers, we further compare representative trackers with different search strategies to evaluate whether global search, re-detection, or adaptive search-region modeling can better handle target loss caused by severe spatial shift. As shown in Table~\ref{tab:global_redetection}, SPLT~\cite{yan2019skimming}, as a re-detection tracker, obtains 37.4\% PR, 32.8\% NPR, and 28.1\% SR. GlobalTrack~\cite{huang2020globaltrack}, as a global-search tracker, achieves 39.8\% PR, 34.2\% NPR, and 30.0\% SR. ZoomTrack obtains stronger performance with 47.1\% PR, 41.6\% NPR, and 35.2\% SR, indicating that adaptive search-region modeling is helpful for handling large target displacement. Nevertheless, SARLA still achieves the best performance with 50.6\% PR, 44.8\% NPR, and 38.0\% SR.}

\textcolor{red}{These results suggest that global search and re-detection strategies can alleviate target loss to some extent, but they remain insufficient for cross-modal UAV object tracking, because modality switch simultaneously causes severe appearance variation and spatial shift. In contrast, SARLA explicitly models the appearance state and spatial state caused by modality switch, leading to more robust cross-modal tracking performance.}

\begin{table}[tbp!]
\caption{Quantitative comparison against excellent trackers. Without retraining indicates that the trackers are not trained with the training set from our dataset, while Retraining signifies that they are trained with it.}\label{tab:retrain}
\centering                  
\begin{tabular}{l|ccc|ccc}
\toprule
\multirow{2}{*}{Method} & ~ & Without retraining & ~ & ~ & Retraining & ~ \\ 
 & PR & NPR & SR & PR & NPR & SR \\ 
\midrule
TransT & 35.1 & 30.2 & 26.3 & 39.5 & 33.6 & 29.2 \\ 
OSTrack & 27.9 & 23.2 & 21.1 & 37.0 & 32.7 & 28.0 \\ 
HiT  & 25.9 & 21.4 & 20.4 & 40.0 & 34.3 & 30.1 \\
SeqTrack & 32.3 & 28.9 & 24.9 & 43.3 & 39.2 & 33.5 \\
GRM & 32.3 & 27.0 & 24.6 & 38.5 & 33.5 & 28.8 \\ 
EVPTrack & 30.6 & 26.4 & 23.9 & 46.5 & 41.7 & 35.7 \\
AVTrack & 30.2 & 24.4 & 22.6 & 38.0 & 32.0 & 28.1 \\
AQATrack & 31.3 & 27.3 & 24.4 & 44.5 & 40.1 & 34.2 \\
\textcolor{red}{SUTrack} & \textcolor{red}{39.1} & \textcolor{red}{35.3} & \textcolor{red}{31.0} & \textcolor{red}{48.0} & \textcolor{red}{42.3} & \textcolor{red}{36.7} \\
\textcolor{red}{SpikeTrack} & \textcolor{red}{32.2} & \textcolor{red}{27.8} & \textcolor{red}{24.7} & \textcolor{red}{45.1} & \textcolor{red}{40.5} & \textcolor{red}{34.8} \\
\midrule
SARLA & - & - & - & \textbf{50.6} & \textbf{44.8} & \textbf{38.0} \\
\bottomrule
\end{tabular}
\end{table}

\begin{table}[tbp!]
\caption{\textcolor{red}{Tracking performance on aligned CM-UOT.}}\label{tab:aligned_cmuot}
\centering
\begin{tabular}{l|ccc|ccc}
\toprule
\multirow{2}{*}{Method} & ~ & Without retraining & ~ & ~ & Retraining & ~ \\
& PR & NPR & SR & PR & NPR & SR \\
\midrule
OSTrack & 27.6 & 23.4 & 21.4 & 45.6 & 41.2 & 35.2 \\
HiT & 26.7 & 21.7 & 20.8 & - & - & - \\
SeqTrack & 36.6 & 32.4 & 27.9 & - & - & - \\
GRM & 39.5 & 33.1 & 29.1 & 52.1 & 45.6 & 39.1 \\
EVPTrack & 34.0 & 29.2 & 26.2 & 48.3 & 44.2 & 37.7 \\
AQATrack & 34.8 & 30.2 & 26.7 & 52.8 & 45.9 & 39.6 \\
ZoomTrack & 34.1 & 29.6 & 26.2 & - & - & - \\
\midrule
SARLA & - & - & - & \textbf{54.8} & \textbf{47.9} & \textbf{41.2} \\
\bottomrule
\end{tabular}
\end{table}

\subsection{Ablation Study}
\noindent
\textbf{Component Ablation.}
The ablation comparisons in Table~\ref{tab:ablation} analyze the effectiveness of each component in our proposed SARLA model.
SARLA seamlessly integrates modality state information and spatial shift state information. To comprehensively evaluate the individual contributions of these two types of information to SARLA’s performance, we conduct a series of ablation experiments by selectively ablating the Modality State Aware Representation Module (MSARM) and the Spatial State Aware Representation Module (SSARM).
The results in Rows 2 and 3 of Table~\ref{tab:ablation} demonstrate that incorporating either MSARM or SSARM alone yields significant performance improvements over the baseline (without these modules). This is because MSARM is specifically designed to address modality discrepancy issues, which is particularly beneficial in scenarios with small spatial offsets; in contrast, SSARM focuses on mitigating the adverse effects of spatial shift, thus enhancing performance when handling large target shift.
Leveraging this complementary property between the two modules, their integration (Row 4 in Table~\ref{tab:ablation}) achieves further performance gains compared to the single-module variants.
As illustrated in Rows 3, 5 and 4, 6 of Table~\ref{tab:ablation}, the proposed spatial shift prediction loss (SSPL) plays a critical role in guiding the model to learn accurate spatial modeling.
Collectively, these results verify that the synergistic integration of MSARM, SSARM, and SSPL enables the model to simultaneously tackle both core challenges (appearance change and spatial shift), leading to more robust and consistent tracking performance.

\begin{table}[htbp]
\caption{Summary of cumulative effects. MST denotes the Modality State Aware Representation Module; SST denotes the Spatial State Aware Representation Module; SSPL denotes the spatial shift prediction loss.}\label{tab:ablation}
\centering
\begin{tabular}{l|ccc|ccc}
\toprule
No. & MSARM  & SSARM & SSPL    & PR   & NPR  & SR   \\ 
\midrule
1 &                            &                     &   & 37.0 & 32.7 & 28.0 \\
2 & \checkmark                 &                     &   & 45.9 & 39.9 & 34.2 \\
3 &                            &   \checkmark        &   & 45.8 & 39.9 & 34.4 \\
4 &  \checkmark                &   \checkmark        &  & 49.4 & 43.6 & 37.2 \\
5 &                            &   \checkmark        & \checkmark  & 47.2 & 41.6 & 35.8 \\
6 & \checkmark                 &    \checkmark       & \checkmark  & \textbf{50.6} & \textbf{44.8} & \textbf{38.0} \\ 
\bottomrule
\end{tabular}
% }
\end{table}

\subsection{\textcolor{red}{Parameter Analysis}}
\begin{figure}[htbp]
\centering
\includegraphics[width=0.75\linewidth]{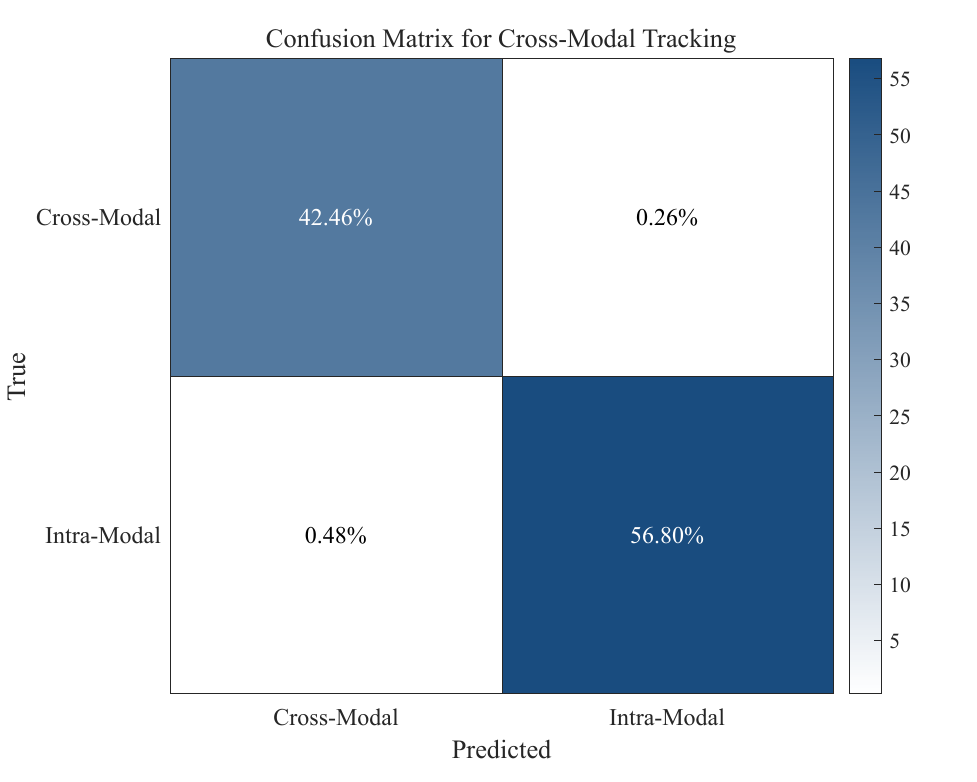}
\caption{\textcolor{red}{Confusion matrix of cross-modal and intra-modal state prediction.}}
\label{fig:confusion_matrix}
\end{figure}

\noindent
\textbf{\textcolor{red}{State Prediction Analysis}.}
\textcolor{red}{To further analyze the state-aware mechanism itself, we report the confusion matrix of the state-token classification results in Fig.~\ref{fig:confusion_matrix}. The confusion matrix provides a direct diagnosis of whether the learned state representation can distinguish cross-modal and intra-modal states. As shown in Fig.~\ref{fig:confusion_matrix}, most samples are correctly classified into their corresponding states, with 42.46\% cross-modal samples and 56.80\% intra-modal samples located on the diagonal. In contrast, only 0.26\% cross-modal samples are misclassified as intra-modal, and only 0.48\% intra-modal samples are misclassified as cross-modal. The overall classification accuracy reaches 99.26\%. These results demonstrate that the learned state-token representation can effectively capture modality-state information.}

\begin{figure}[htbp]
\centering
\includegraphics[width=0.95\linewidth]{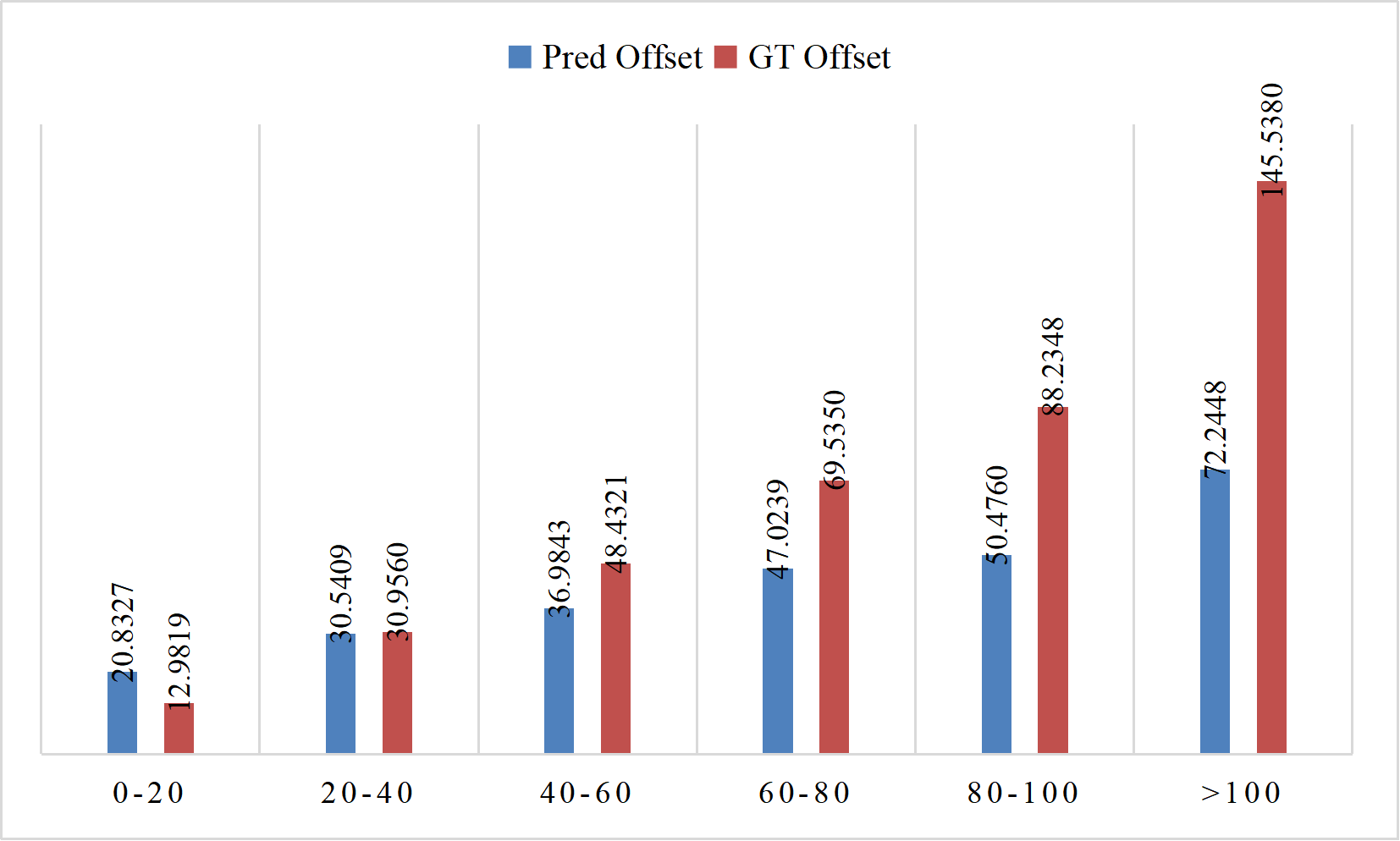}
\caption{\textcolor{red}{Comparison between predicted offset and ground-truth offset across different shift ranges on CM-UOT.}}
\label{fig:offset_pred}
\end{figure}

\bigskip\noindent
\textbf{\textcolor{red}{Spatial-Shift Prediction Analysis}.}
\textcolor{red}{We further divide the samples into different groups according to the ground-truth target-center offset, and compare the mean predicted offset with the mean ground-truth offset in each group. As shown in Fig.~\ref{fig:offset_pred}, the predicted offset generally increases as the ground-truth offset range becomes larger. This indicates that the shift prediction head is not making random predictions, but can capture the overall tendency of spatial displacement. However, in large-offset cases, the prediction is still not sufficiently accurate, especially because the predicted offset is clearly smaller than the ground-truth offset. Therefore, the predicted shift is more suitable as auxiliary supervision during training to help the model learn spatial-aware representations, rather than being directly used to correct the final tracking result during inference.}

\noindent
\textbf{\textcolor{red}{Evaluation on Aligned CM-UOT}.}
\textcolor{red}{We further evaluate representative trackers on aligned CM-UOT to disentangle the effects of spatial shift and cross-modal appearance discrepancy. Since aligned CM-UOT removes the spatial displacement caused by modality switching while preserving the modality gap, the performance difference between CM-UOT and aligned CM-UOT reflects the influence of spatial shift. As shown in Table~\ref{tab:aligned_cmuot}, most trackers perform better on aligned CM-UOT after retraining. For example, SARLA improves from 50.6\% PR, 44.8\% NPR, and 38.0\% SR to 54.8\% PR, 47.9\% NPR, and 41.2\% SR, indicating that spatial shift is an important factor limiting cross-modal UAV tracking.}

\textcolor{red}{Nevertheless, the performance on aligned CM-UOT is still far from saturated, suggesting that cross-modal appearance discrepancy remains challenging even after removing spatial shift. We also include ZoomTrack as a representative tracker with stronger search ability. It achieves 34.1\% PR, 29.6\% NPR, and 26.2\% SR under without retraining on aligned CM-UOT, showing that enlarging the search range alone is insufficient under large modality gaps. These results further demonstrate the necessity of SARLA's modality-state and spatial-state modeling.}

\noindent
\textbf{\textcolor{red}{State and Shift Prediction Analysis}.}
\textcolor{red}{We further analyze two design choices related to state-aware representation learning and spatial-shift prediction. First, to investigate the necessity of the proposed state-token mechanism, we compare SARLA with a variant that replaces the learnable state tokens with pooled search-region features. As shown in Table~\ref{tab:state_shift_analysis}, directly using search-region features only obtains 44.9\% PR, 40.2\% NPR, and 34.5\% SR, which is clearly lower than SARLA. This indicates that raw search-region features cannot effectively capture the modality-state and spatial-state variations required for robust cross-modal UAV tracking. Second, we evaluate whether the predicted spatial shift should be explicitly used during inference. Using the predicted shift to guide the Hanning window does not improve performance, suggesting that directly applying an inaccurate shift prediction may introduce additional uncertainty. Therefore, SARLA uses the spatial shift prediction as an auxiliary training objective to enhance the spatial-aware representation, rather than as an explicit inference-time correction.}

\begin{table}[htbp]
\centering
\caption{\textcolor{red}{Analysis of state-token mechanism and inference-time shift usage on CM-UOT.}}
\label{tab:state_shift_analysis}
\setlength{\tabcolsep}{4pt}
\begin{tabular}{lccc}
\toprule
Setting & PR & NPR & SR \\
\midrule
SARLA & 50.6 & 44.8 & 38.0 \\
w/o State Tokens & 44.9 & 40.2 & 34.5 \\
w/ Use of Predicted Offset & 50.2 & 44.5 & 37.8 \\
\bottomrule
\end{tabular}
\end{table}

\noindent
\textbf{\textcolor{red}{Efficiency Analysis}.}
\textcolor{red}{
To further support the efficiency of the proposed single-stream design, we report FPS, FLOPs, testing parameters, and tracking performance in Table~\ref{tab:efficiency}. Compared with OSTrack, SARLA keeps the same testing parameters (90.1M) and maintains comparable real-time speed, with FPS decreasing slightly from 93.1 to 89.6, while improving PR/NPR/SR from 37.0\%/32.7\%/28.0\% to 50.6\%/44.8\%/38.0\%. Compared with OSTrack-TwoBranch, SARLA achieves lower FLOPs (51.1G vs. 58.7G), faster inference speed (89.6 vs. 67.1 FPS), and better tracking accuracy. These results verify that SARLA provides a better accuracy-efficiency trade-off within a unified single-stream framework.
}
% \begin{table}[htbp]
% \centering
% \caption{\textcolor{red}{Efficiency comparison between OSTrack and SARLA.}}
% \label{tab:efficiency}
% % \setlength{\tabcolsep}{4pt}
% \begin{tabular}{lcccccc}
% \toprule
% Method & FPS &  Flops(G) &  Testing Params(M) & PR & NPR & SR  \\
% \midrule
% OSTrack & 93.1 & & 90.1 & 37.0 & 32.7 & 28.0 \\
% OSTrack-TwoBranch & & &  & 38.6 & 33.1 & 28.6 \\
% SARLA & 89.6 & & 90.1 & \textbf{50.6} & \textbf{44.8} & \textbf{38.0} \\
% \bottomrule
% \end{tabular}
% \end{table}

\begin{table}[htbp]
\centering
\caption{\textcolor{red}{Efficiency and accuracy comparison on CM-UOT.}}
\label{tab:efficiency}
\setlength{\tabcolsep}{3.5pt}
\begin{tabular}{lcccccc}
\toprule
Method & FPS & \makecell{FLOPs\\(G)} & \makecell{Testing\\Params (M)} & PR & NPR & SR \\
\midrule
OSTrack & 93.1 & 29.0 & 90.1 & 37.0 & 32.7 & 28.0 \\
\makecell[l]{OSTrack-\\TwoBranch} & 67.1 & 58.7 & 91.3 & 38.6 & 33.1 & 28.6 \\
SARLA & 89.6 & 51.1 & 90.1 & \textbf{50.6} & \textbf{44.8} & \textbf{38.0} \\
\bottomrule
\end{tabular}
\end{table}

\begin{figure*}[tbp!]
\centering
\includegraphics[width=0.8\linewidth]{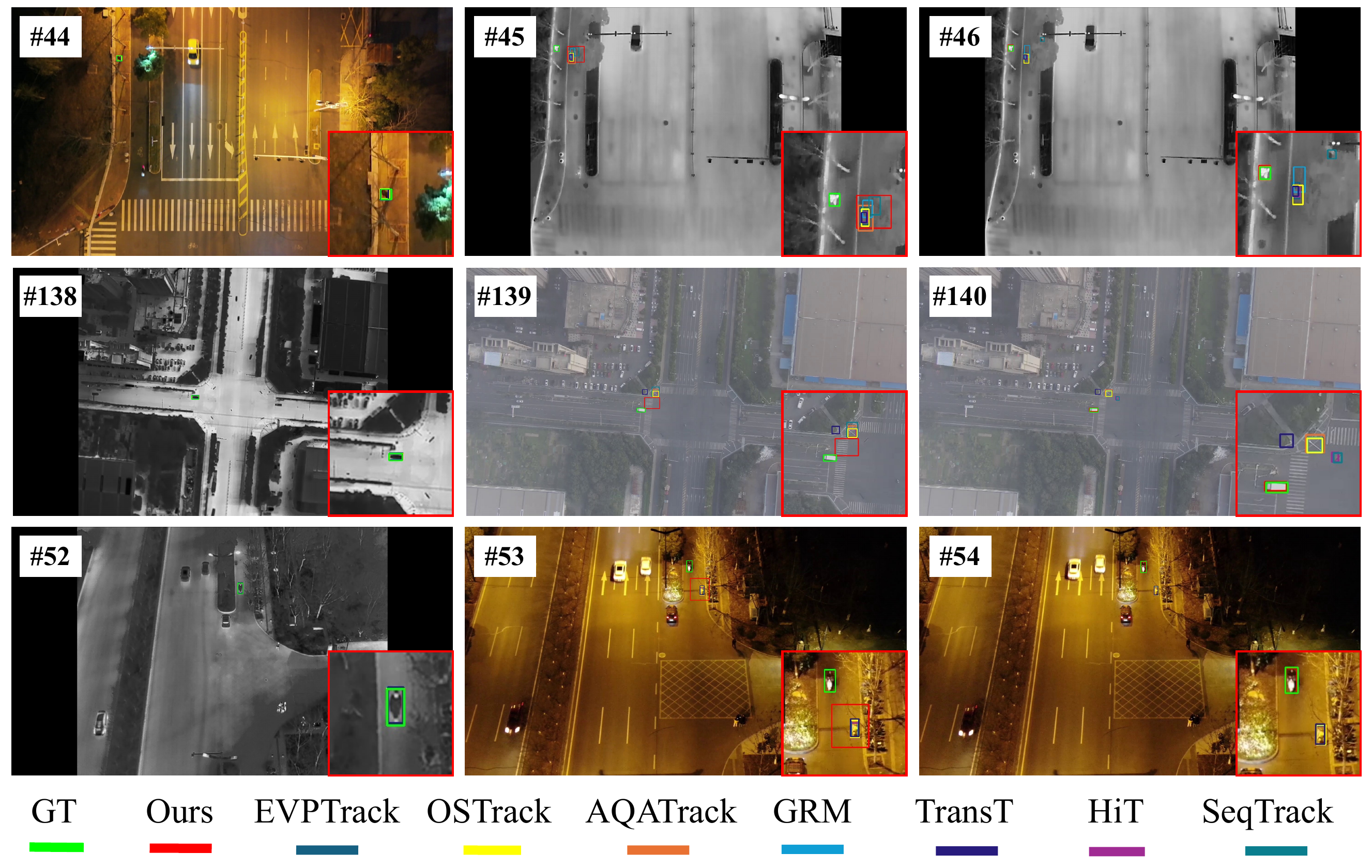}
\caption{Visual comparison on three representative sequences. The purple box at the bottom right is a locally enlarged view for clearer visualization of the tracking results.}
\label{fig:vis}
\end{figure*}

\subsection{Qualitative Comparison}  
As shown in Fig.~\ref{fig:vis}, we present qualitative tracking results between our method and four other excellent trackers.
Three representative sequences, which include significant appearance change and sudden spatial shift during modality switch, are selected from the CM-UOT benchmark to compare the performances of different methods. Before the modality switch, all trackers perform well and find the position of the target. After the switch, it is not hard to see that the extent of the spatial shift is so large that all trackers fail to locate the target object and instead locate artifacts. However, our tracker takes some actions. As we can see in the second column of the pictures in Fig.~\ref{fig:vis}, the prediction of our tracker is much larger than that of others. In the subsequent frame, the target is relocated, whereas other trackers mistakenly adhere to incorrect results. 

Since the size of the search region is fixed and the cropping strategy is based on the previously predicted center position, its size being determined by multiplying the previously predicted size by a fixed factor and then resizing it to the fixed size, a larger prediction implies that the search region will encompass a broader scope of the frame in the subsequent frame. This is conducive to scenarios where the offset distance is considerably large or even when the target is outside the search region.
Moreover, this phenomenon persists for an extended period, underscoring the urgency of addressing the challenges posed by significant appearance change and sudden spatial shift.
These cases demonstrate the capability of our algorithm to effectively adapt to cross-modal appearance change and large spatial shift, thereby validating the effectiveness of state-aware representation learning.

\subsection{Evaluation on CMOTB}
To further validate the effectiveness of our SARLA method, we conduct experiments on the CMOTB dataset. As presented in Table~\ref{tab:cmotb}, SARLA, MAFNet$_{DiMP}$, and OSTrack rank first, second, and third, respectively. 
Compared to DiMP, MAFNet$_{DiMP}$ incorporates adaptive modality-aware fusion to bridge the modality gap, achieving enhancements of 4.0\% in PR, 4.0\% in NPR, and 3.8\% in SR. However, MAFNet$_{DiMP}$, being CNN-based, has limitations in global context modeling. 
In contrast, SARLA, similar to OSTrack, which performs feature extraction and relation modeling through a self-attention operation, attains better performance than MAFNet$_{DiMP}$.

\begin{table}[htbp]
\caption{Tracking results of trackers in the testing set of CMOTB dataset.}\label{tab:cmotb}
\centering
\begin{tabular}{c|ccc}
\toprule
Method & PR & NPR & SR \\ 
\midrule
MAFNet$_{RT}$ & 53.0 & 56.2 & 43.7 \\
DiMP & 70.4 & 73.8 & 60.0 \\ 
MAFNet$_{DiMP}$ & 74.4 & 77.8 & 63.8 \\
TrDiMP & 66.8 & 69.9 & 58.3 \\
TransT & 68.9 & 72.0 & 59.1 \\
OSTrack & 74.3 & 76.8 & 63.3 \\ 
\midrule
SARLA & \textbf{76.6} & \textbf{79.0} & \textbf{65.1} \\
\bottomrule
\end{tabular}
\end{table}

\subsection{Limitation}
SARLA is capable of handling spatial shift challenges to a certain extent. Nevertheless, when the degree of spatial shift is substantial, for instance, when the target is located outside the search region, our method is unable to handle such situations within the frame during modality switch. 
We discuss potential remedies for these issues. 
Most trackers rely on cropped regions for target tracking and struggle to handle the situation where the target moves out of the cropped region. 
Therefore, it is necessary to construct a network to ensure that the target remains within the search region. This network will utilize a larger region or even the entire frame to help re-crop a more favorable search region.

\section{Conclusion}
In this paper, we propose a State-Aware Representation Learning Approach for cross-modal UAV object tracking, which model cross-modality appearance correlation and cross-frame spatial correlation to adapt to sudden changes in both appearance and position.
Furthermore, to foster research and development within this field, we pioneered the establishment of a large-scale video benchmark. This benchmark encompasses both unaligned and aligned versions, taking into account the spatial shift issue that arises from modality switch.
Experiments conducted on this benchmark convincingly demonstrate that our method significantly outperforms existing superior trackers.
In the future, we will refine our method to handle scenarios where modality switch accompanies large spatial shift. 
Additionally, we will further explore plug-and-play solutions to address substantial appearance change and sudden spatial shift, facilitating flexible integration into diverse tracking frameworks.

\bibliographystyle{IEEEtran}
\bibliography{main}


% \end{IEEEbiography}

\vspace{11pt}

\begin{IEEEbiographynophoto}{Yun Xiao}
received the M.S. and Ph.D. degrees from the School of Computer Science and Technology, Anhui University, Hefei, China, in 2011 and 2019, respectively. She has been a Visiting Student with the Department of Mathematics and Computational Science, University of Stirling, Stirling, U.K., since 2018. She is currently an Associate Professor with the School of Artificial Intelligence, Anhui University. Her research interests include computer vision and deep learning.
\end{IEEEbiographynophoto}

\begin{IEEEbiographynophoto}{Zhihong Hong}
    received the B.E. degree in software engineering from Nanchang University, Nanchang, China. He is currently pursuing the M.S. degree with Anhui University, Hefei, China, in 2019.
    His current research interests include computer vision and visual object tracking.
\end{IEEEbiographynophoto}

\begin{IEEEbiographynophoto}{Jiandong Jin}
    received the B.E. degree in communication engineering from Anhui Polytechnic University, Wuhu, China. He is currently pursuing the M.S. degree with Anhui University, Hefei, China.
    His current research interests include computer vision and pedestrian attribute recognition.
\end{IEEEbiographynophoto}

\begin{IEEEbiographynophoto}{Chenglong Li}
 received the M.S. and Ph.D. degrees from the School of Computer Science and Technology, Anhui University, Hefei, China, in 2013 and 2016, respectively. From 2014 to 2015, he was a Visiting Student with the School of Artificial Intelligence, Sun Yat-sen University, Guangzhou, China. He was a Post-Doctoral Research Fellow with the Center for Research on Intelligent Perception and Computing (CRIPAC), National Laboratory of Pattern Recognition (NLPR), Institute of Automation, Chinese Academy of Sciences (CASIA), Beijing, China. He is currently a Professor and a Ph.D. Supervisor with the School of Artificial Intelligence, Anhui University. His research interests include computer vision and deep learning.
\end{IEEEbiographynophoto}

\begin{IEEEbiographynophoto}{Jin Tang}
 received the B.E. degree in automation and the Ph.D. degree in computer science from Anhui University, Hefei, China, in 1999 and 2007, respectively.,He is currently a Professor with the School of Computer Science and Technology, Anhui University. His research interests include computer vision, pattern recognition, machine learning, and deep learning.
\end{IEEEbiographynophoto}

\begin{IEEEbiographynophoto}{Amir Hussain}
 received the B.Eng. (Hons.) and Ph.D. degrees in electronic and electrical engineering from the University of Strathclyde, Glasgow, U.K., in 1992 and 1997, respectively.,He is Founding Director of the Centre of AI and Robotics with Edinburgh Napier University, Edinburgh, U.K. He has led major national and international projects and supervised over 40 Ph.D. students. He has authored several patents and more than 600 publications, including around 300 journal papers and 20 books/monographs. His research interests are cross-disciplinary and industry-led, and aimed at developing responsible AI and cognitive data science technologies to engineer the smart healthcare and industrial systems of tomorrow.,Dr. Hussain is Founding Chief Editor of Springer’s Cognitive Computation journal and Invited Editor for various other journals. Amongst other distinguished roles, he has served as General Chair of the flagship 2020 IEEE World Congress on Computational Intelligence and the 2023 IEEE Smart World Congress He is Chair of the IEEE U.K. and Ireland Chapter of the IEEE Industry Applications Society.
\end{IEEEbiographynophoto}

\vfill

\end{document}